\documentclass[lettersize,journal]{IEEEtran}
\usepackage{amsmath,amsfonts}
\usepackage{algorithmic}
\usepackage{algorithm}
\usepackage{array}
\usepackage[caption=false,font=normalsize,labelfont=sf,textfont=sf]{subfig}
\usepackage{textcomp}
\usepackage{stfloats}
\usepackage{url}
\usepackage{verbatim}
\usepackage{graphicx}
\usepackage{cite}
\usepackage{bbm}
\usepackage{multirow}
\usepackage{booktabs}
\usepackage[citecolor=blue, colorlinks]{hyperref}
\usepackage{tikz}
\usetikzlibrary{backgrounds,automata}

\begin{document}

\title{CorrI2P: Deep Image-to-Point Cloud Registration via Dense Correspondence}

\author{Siyu Ren, Yiming Zeng, Junhui Hou,~\IEEEmembership{Senior Member,~IEEE,} Xiaodong Chen
\thanks{This project was supported in part by the Hong Kong Research Grants Council under Grants 11202320 and 11218121, and in part by the Natural Science Foundation of China under Grant 61871342.  \textit{Corresponding author: Junhui Hou}}
\thanks{S. Ren, Y. Zeng, and J. Hou are with the Department of Computer Science,
City University of Hong Kong, Hong Kong (email: \href{siyuren2-c@my.cityu.edu.hk}{siyuren2-c@my.cityu.edu.hk}; \href{ymzeng4-c@my.cityu.edu.hk}{ymzeng4-c@my.cityu.edu.hk}; \href{jh.hou@cityu.edu.hk}{jh.hou@cityu.edu.hk}). }
\thanks{X. Chen is with the School of Precision Instrument and Opto-electronics Engineering, Tianjin University, Tianjin, China (email: \href{xdchen@tju.edu.cn}{xdchen@tju.edu.cn}).}}

\markboth{}%
{Shell \MakeLowercase{\textit{et al.}}: A Sample Article Using IEEEtran.cls for IEEE Journals}


\maketitle

\begin{abstract}
Motivated by the intuition that the critical step of localizing a 2D image in the corresponding 3D point cloud is establishing 2D-3D correspondence between them, we propose the first feature-based dense correspondence framework for addressing the challenging problem of 2D image-to-3D point cloud registration, dubbed CorrI2P. CorrI2P is mainly composed of three modules, i.e., feature embedding, symmetric overlapping region detection, and pose estimation through the established correspondence. Specifically, given a pair of a 2D image and a 3D point cloud, we first transform them into high-dimensional feature spaces and feed the resulting features into a symmetric overlapping region detector to determine the region where the image and point cloud overlap. Then we use the features of the overlapping regions to establish dense 2D-3D correspondence, on which EPnP within RANSAC is performed to estimate the camera pose, i.e., translation and rotation matrices. Experimental results on KITTI and NuScenes datasets show that our CorrI2P outperforms state-of-the-art image-to-point cloud registration methods significantly. The code will be publicly available at \href{https://github.com/rsy6318/CorrI2P}{https://github.com/rsy6318/CorrI2P}.
\end{abstract}


\section{Introduction}
\IEEEPARstart{V}{isual} pose estimation is a critical task for autopilot \cite{AUTOPILOT}, robotics \cite{ROBOTICS}, and augmented/mixed reality \cite{AR} devices since it is the base of SLAM \cite{SLAM2,SLAM} and Structure-from-Motion \cite{SFM}. Its objective is to determine the image's 6DOF camera position in a 3D scene. Finding the correspondence between them is the key step in this procedure, followed by utilizing algorithms like EPnP \cite{EPNP} to generate the camera posture based on the relationship.

The registration problem is closely related to visual pose estimation depending on the establishment of correspondence. According to the modality of data, it can be divided into two categories, same-modal and cross-modal registration. Many solutions for same-modal registration have been presented, such as image-to-image (I2I) \cite{I2I1,I2I2,I2I3,TCSVTI2I,TCSVTI2I2} or point cloud-to-point cloud (P2P) registration \cite{ICP,FGR,GOICP,3DMATCH,PERFECTMATCH,PREDATOR,DEEPI2P,DEEPICP,TCSVTP2P,TCSVTP2P2,TCSVTP2P3,CORRNET3D,feng2021recurrent}. 
Specifically, I2I registration builds 2D-2D correspondence between images, but it only uses color information and could be influenced by the environment. P2P registration builds 3D-3D correspondence between point clouds, and thus, it needs large storage space. Both 2D-2D and 3D-3D correspondences are same-modal. Cross-modal data registration, i.e., image-to-point cloud (I2P) registration \cite{2D3DMATCHNET,DEEPI2P}, can remedy the disadvantages of these two same-modal techniques. However, it relies on cross-modal correspondence, i.e., 2D-3D correspondence, which is more challenging to estimate.
Previous works for I2I and P2P registration 
cannot be simply extended to establishing 2D-3D correspondence in I2P registration because they establish 2D-2D or 3D-3D correspondence from same-modal data. SfM \cite{SFM} is a well-known approach for obtaining 2D-3D correspondence, which reconstructs 3D point clouds from a series of images and obtains correspondence based on image features. 
However, the reconstructed point cloud from images has low accuracy and suffers from the limitation that image features are easily influenced by external environments.
2D3D-MatchNet \cite{2D3DMATCHNET} is the first feature-based registration method, which seeks  2D-3D correspondence directly. However, it focuses on the correspondence of key points detected based on the hand-crafted features by SIFT \cite{SIFT} and ISS \cite{ISS}. The above-mentioned methods are feature-based, meaning that these methods use features of pixels and points to establish 2D-3D correspondence according to the nearest neighborhood principle. 
The recent DeepI2P \cite{DEEPI2P} casts the correspondence problem to point-wise classification without utilizing 
pixel-wise or point-wise features to establish 2D-3D correspondence. 
However, the points at the frustum boundary are prone to be wrongly classified, 
thus limiting registration accuracy.

In this paper, we propose CorrI2P, the first learnable paradigm for building \textit{dense} 2D-3D correspondence. It is a two-branch neural network with a symmetric cross-attention fusion module identifying overlap and extracting dense features from the image and point cloud. Based on the extracted overlap features, it builds the 2D-3D correspondence and estimates the camera's posture. We design a descriptor loss and a detector loss to drive the training of CorrI2P. Experimental results show that our CorrI2P achieves state-of-the-art performance on KITTI \cite{KITTI} and NuScenes \cite{NUSCENES} datasets.

In summary, the main contributions of this paper are three-fold:
\begin{enumerate}
\item we propose the {first} feature-based {dense} correspondence framework for image-to-point cloud registration;
\item we design a novel {symmetric overlapping region detector} for the cross-modal data, i.e., images and point clouds; and
\item we propose a joint loss function to drive the learning process of the cross-modal overlap detection and dense correspondence.
\end{enumerate}

The rest of this paper is organized as follows. Sec. \ref{sec:RW} briefly reviews existing works about visual pose estimation and 2D-3D correspondence for registration. Sec. \ref{sec:proposed} presents the proposed CorrI2P, followed by extensive experiments and analysis in Sec. \ref{sec:exp}. Finally, Sec. \ref{sec:con} concludes this paper.  

\section{Related Works}
\label{sec:RW}
\subsection{Visual Pose Estimation}
Given a query image, estimating the 6DOF camera pose in a 3D scene model, usually presented as a point cloud, is crucial for visual navigation. The critical step in this progress is to build 2D-3D correspondence between the image and point cloud. SfM \cite{SFM} is a traditional method of recovering the point cloud from a sequence of images while using the handcrafted image feature (SIFT \cite{SIFT}, ORB \cite{ORB}, or FAST \cite{FAST}) to generate the 2D-3D correspondence. It utilizes pixel-wise features and the recovered 3D points to establish the 2D-3D correspondence. However, the reconstructed point cloud is sparse, and the imaging environment can affect pixel-wise features.

Some learnable approaches based on same-modal data, i.e., I2I or P2P, have been proposed with the emergence of deep learning. As for the I2I registration methods, \cite{overtime1,overtime2} collect the images from different environments and train CNNs to extract robust features to establish correspondence. Furthermore, \cite{REGPOSE1,REGPOSE2,REGPOSE3} use CNNs to regress camera pose directly. They benefit from the easy availability of image data but are susceptible to environmental conditions. P2P registration methods, on the other hand, obtain accurate point cloud data from Lidar or RGBD cameras.

With the growth of deep learning on the point cloud, \cite{DEEPICP,DCP,RPMNET,3DMATCH,D3FEAT,POINTNETLK,PCRNET,PERFECTMATCH,PREDATOR} use neural networks to extract point-wise or global features from the point cloud directly and combine some traditional methods, such as RPM \cite{RPM} and RANSAC \cite{RANSAC}, to estimate the rigid transformation. The above same-modal methods cannot be readily extended to I2P registration, which relies on learned 2D-3D correspondence from cross-modal data.

\begin{figure*}[t]
\centering
\includegraphics[width=1\textwidth]{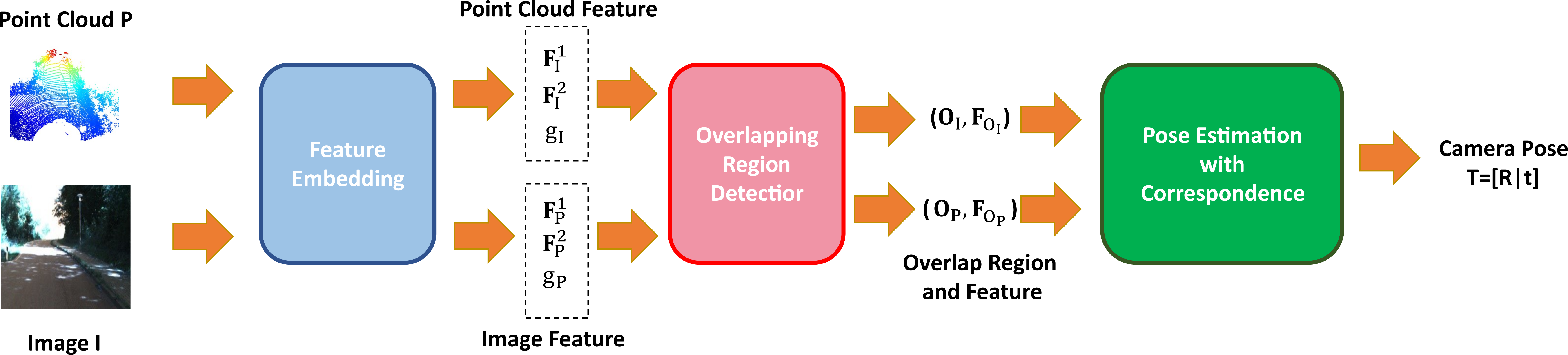}
\caption{Illustration of the overall pipeline of the proposed CorrI2P for image-to-point registration. Taking a pair of a 2D RGB image and a 3D point cloud as input, CorrI2P first performs feature embedding, producing pixel-wise and point-wise features of two scales, as well as global features. Then the resulting features are fed into an overlapping region detector to detect the overlapping regions on both the image and point cloud, and the features of pixels and points located in the detected overlapping regions are adopted to 
establish 2D-3D correspondence. Finally, the camera pose can be obtained by applying EPnP within RANSAC on the dense correspondence.}
\label{FIG2}
\end{figure*}

\subsection{2D-3D Correspondence for Registration}
Unlike 2D images represented with regular dense grids, 3D point clouds are irregular and unordered,  posing substantial challenges to learning correspondence between these two modalities. 2D3D-MatchNet \cite{2D3DMATCHNET} selects key points from the image and point cloud using SIFT \cite{SIFT} and ISS \cite{ISS}, respectively, then feeds the patches around these pixels and points to CNNs \cite{VGG} and PointNet \cite{POINTNET} to extract features for creating 2D-3D correspondence. However, independent key point detection for different modalities will reduce the inlier ratio and registration accuracy. DeepI2P \cite{DEEPI2P} employs a feature-free technique in which a network is trained to classify whether each point of the point cloud is located in the visual frustum, i.e., the area where the points could be projected on the image, then inverse camera projection is used to optimize the camera pose until the points identified in the image fall within the frustum. 
However, the points near the border of the frustum are easy to get erroneous classification results. Besides, DeepI2P also divides the image into square regions and uses the network to classify which region the points of the point could be projected onto. The classification result only indicates coarse 2D-3D correspondence, thus limiting the low registration accuracy. 

Yu \textit{et al}. \cite{2D3DCORR} utilized the 2D-3D correspondence established from the linear shapes to estimate the camera pose. But, it requires accurate initialization, which is different from ours. Liu \textit{et al}. \cite{2D3DCORR2} projected the point cloud onto the image and built the correspondence between the projected points and the point cloud. Their 2D-3D correspondence is different from ours, where the image is taken from the camera directly. Agarwal \textit{et al}. \cite{SFM}, Chen \textit{et al}. \cite{overtime1} and Mulfellner \textit{et al}. \cite{overtime2} reconstructed the point cloud from a series of images at different locations. During the reconstruction, they used 2D-3D correspondence to localize each image, but the point cloud features are from the image not the point cloud itself.


\section{Proposed Method}
\label{sec:proposed}


Let $\mathbf{I} \in \mathbb{R}^{W\times H\times 3}$ and $\mathbf{P} \in \mathbb{R}^{N\times 3}$ be a pair of a 2D image and a 3D point cloud, where $W$ and $H$ are the width and height of the image, respectively, and $N$ is the number of points. The objective of I2P registration is to estimate the camera pose in the space of $\mathbf{P}$, denoted as $\mathbf{T} =[\mathbf{R} |\mathbf{t} ]$ with $\mathbf{R} \in \mathbf{SO(3)}$ and $\mathbf{t} \in \mathbb{R}^{3}$, given a query image $\mathbf{I}$. 

As illustrated in Fig. \ref{FIG2}, our method consists of three modules, i.e., feature embedding, symmetric overlapping region detection, and pose estimation with dense correspondence. Specifically, given $\mathbf{I}$ and $\mathbf{P}$, we first embed them into high-dimensional feature spaces separately, then feed the resulting features into the symmetric overlapping region detector to predict the overlapping region and build the dense 2D-3D correspondence, on which we finally run EPnP \cite{EPNP} with RANSAC \cite{RANSAC} to estimate the camera pose. In what follows, we will provide the detail of each module.

\begin{figure*}
\includegraphics[width=0.9\textwidth]{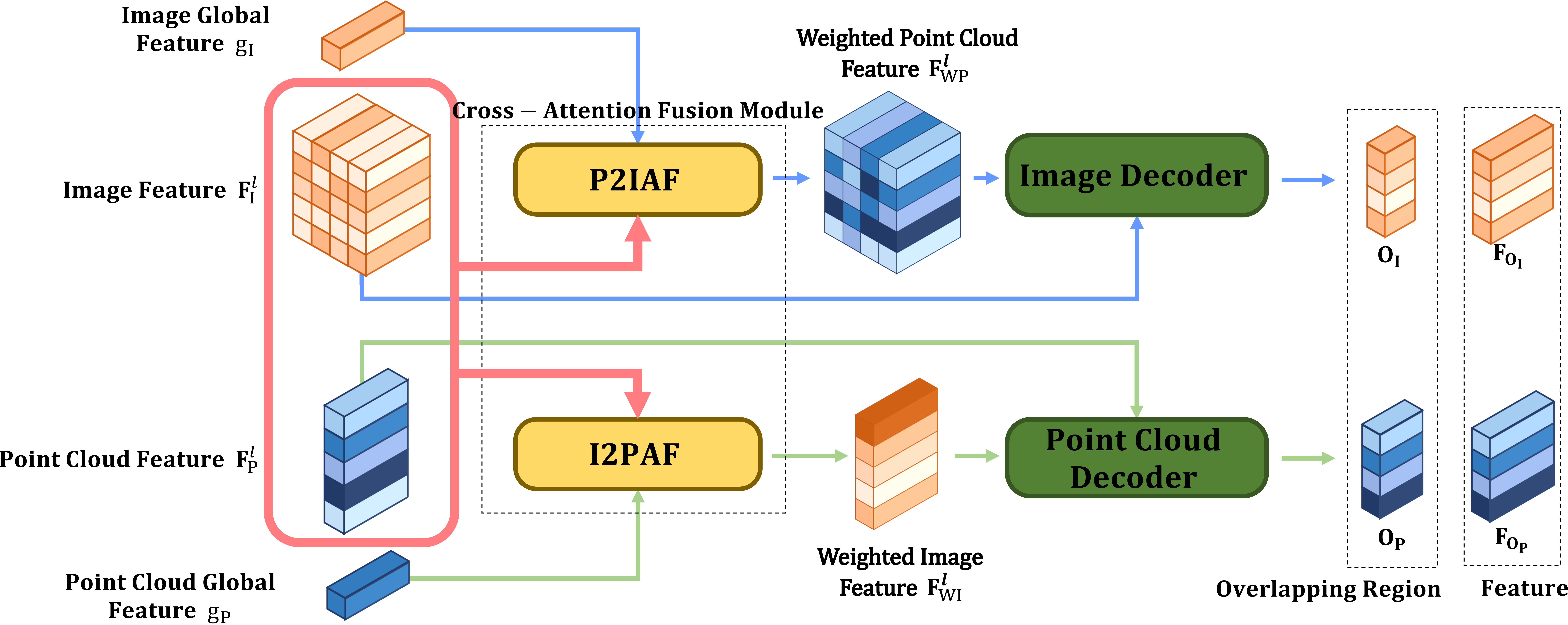}
\caption{Illustration of the network architecture of the symmetric overlapping region detector. 
It is intended to realize the interaction between the features of the image and the point cloud. We feed the images and point cloud features of various scales and their global features into the cross-attention fusion module to map the image and point cloud features into each other's space.  Two decoders then fuse the original and mapped features to determine the overlapping regions.}

\label{FIG3}
\end{figure*}

\begin{figure*}
\centering
\subfloat[]{\label{FIG4A}
\includegraphics[width=0.5\textwidth]{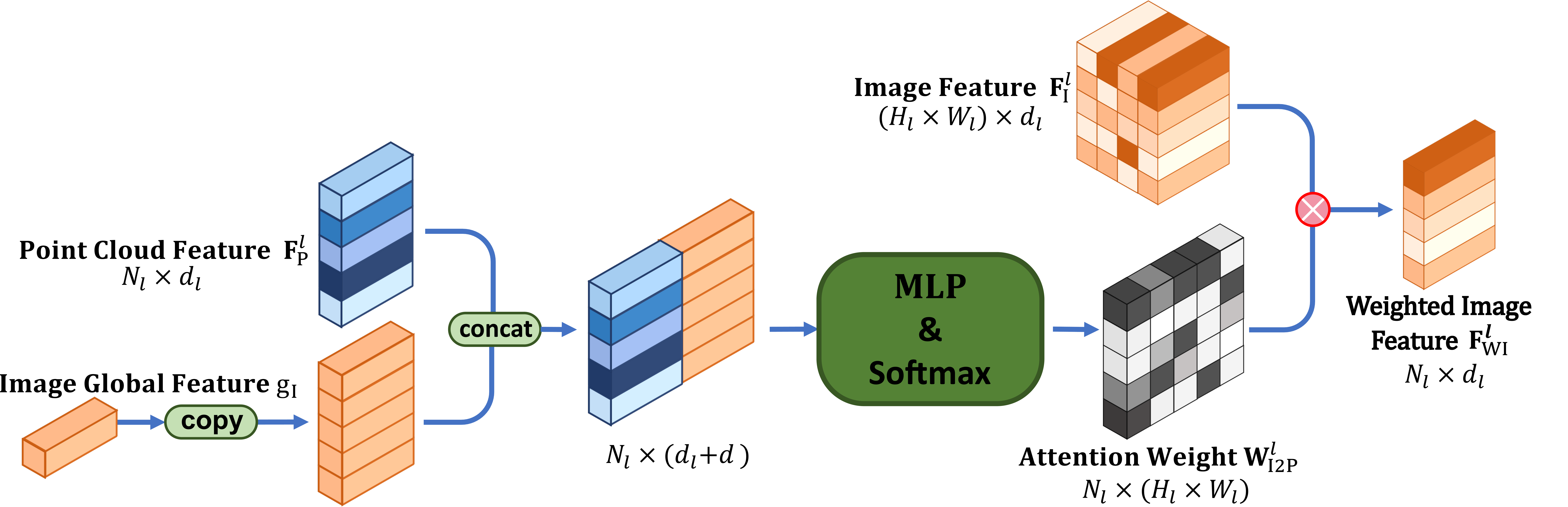}}
\subfloat[]{\label{FIG4B}
\includegraphics[width=0.5\textwidth]{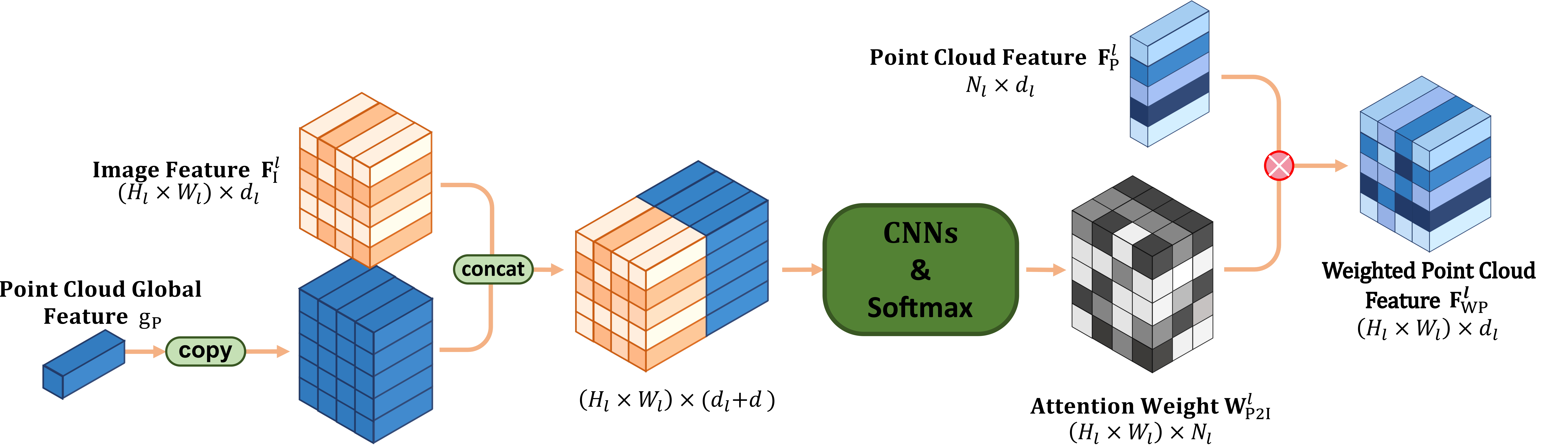}}
\caption{Illustration of the network architectures of the proposed cross-attention fusion module. (a) Image-to-Point Cloud Attention Fusion Module (I2PAF) (b)Point Cloud-to-Image Attention Fusion Module (P2IAF). For I2PAF (resp. P2IAF), the point cloud (resp. image) feature is concatenated with the image (resp. point cloud) global feature and fed into an MLP (resp. a CNN), followed by softmax to generate the attention weight. Then the attention weight is applied to the image (resp. point cloud) feature of the same hierarchy to generate the weighted image (resp. point cloud) feature.} 

\label{FIG4}
\end{figure*}

\subsection{Feature Embedding}
Due to the different structures and characteristics of 2D images and 3D point clouds, it is impossible to deal with them using the same network architecture. Thus, following DeepI2P \cite{DEEPI2P}, we utilize ResNet \cite{RESNet} and SO-Net \cite{SONET} to embed $\mathbf{I}$ and $\mathbf{P}$ to high-dimensional feature spaces in a hierarchical manner, respectively, generating the pixel-wise feature embedding  $\mathbf{F}_{\text{I}}^{l}\in \mathbb{R}^{W_{l} \times H_{l} \times d_{l}}$ and the point-wise feature embedding $\mathbf{F}_{\text{P}}^{l}\in \mathbb{R}^{N_{l} \times d_{l}}$ at the $l$-th layer ($l=1,~2$). Then we perform the max-pooling operation on features $\mathbf{F}_{\text{I}}^{2}$ and $\mathbf{F}_{\text{P}}^{2}$ to obtain the global features of $\mathbf{I}$ and $\mathbf{P}$, denoted as $\mathbf{g}_{\text{I}}\in \mathbb{R}^{d}$ and $\mathbf{g}_{\text{P}}\in \mathbb{R}^{d}$, respectively.

\subsection{Symmetric Overlapping Region Detection}
We design a novel symmetric detector to select the overlapping pixels and points where 2D-3D correspondence is built. As shown in Fig. \ref{FIG3}, we first pass the 2D-3D features into a cross-attention fusion module, composed of two components, namely image-to-point cloud attention fusion (I2PAF) and point cloud-to-image attention fusion (P2IAF), generating weighted features of the image and point cloud. The underlying intuition is to map the image and point cloud features to each other's space. Then we feed the weighted features with the 2D-3D features into the image and point cloud decoders to predict the overlapping regions. 
\\

\noindent\textbf{Cross-attention fusion module}. As shown in Fig. \ref{FIG4}, this module, which aims to fuse the image and point cloud information to detect the overlapping regions, 
consists of I2PAF and P2IAF that are symmetric share a similar structure. 

For I2PAF, we concatenate the image global feature $\mathbf{g}_{\text{I}}$ and point cloud local feature $\mathbf{F}_{\text{P}}^{l}$ and feed them into an MLP followed by the Softmax operator to learn the attention weight $\mathbf{W}_{\text{I2P}}^{l} \in \mathbb{R}^{N_{l} \times (H_{l} \times W_{l} )}$. Then we multiply the image local features $\mathbf{F}_{\text{I}}^{l}$ by attention weights $\mathbf{W}_{\text{I2P}}^{l}$, producing the weighted image feature $\mathbf{F}_{\text{WI}}^{l} \in \mathbb{R}^{N_{l} \times d_{l}}$. Similar to I2PAF, we can get the weighted point cloud feature $\mathbf{F}_{\text{WP}}^{l} \in \mathbb{R}^{H_{l} \times W_{l} \times d_{l}}$ using the symmetric module P2IAF.\\

\begin{figure*}[!htbp]
\centering
\subfloat[]{\label{FIG5A}
\includegraphics[width=1\textwidth]{imgdecoder.pdf}}

\subfloat[]{\label{FIG5B}
\includegraphics[width=1\textwidth]{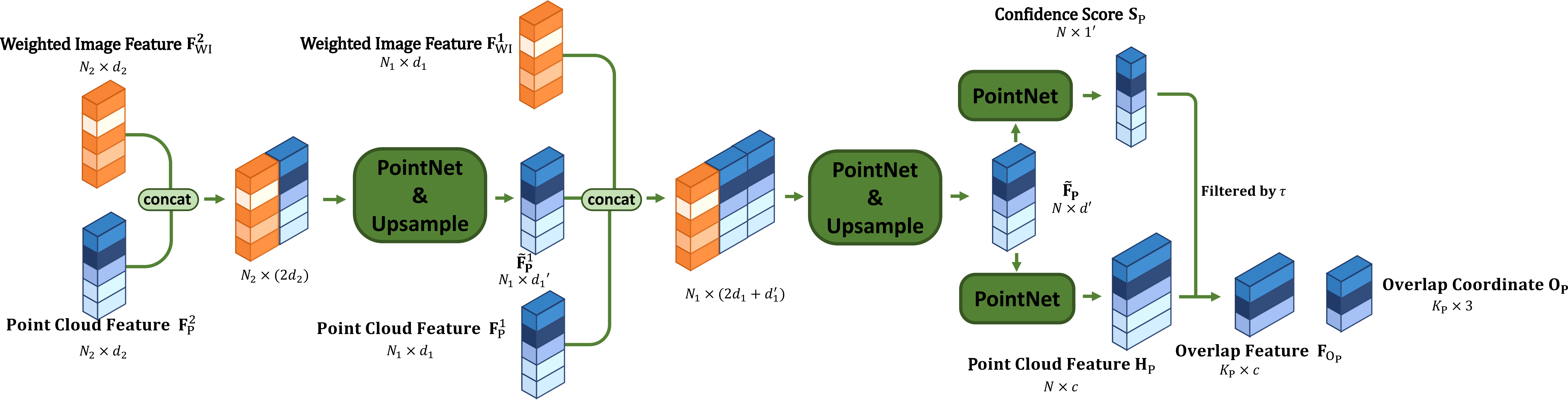}}
\caption{Illustration of the network architectures of the decoder module. The features of the image and point cloud are fed into two detectors, and the coordinates and features of the overlapping regions are produced. (a) Image Decoder. (b) Point Cloud Decoder.}
\label{FIG5}
\end{figure*}

\noindent\textbf{Image and point cloud decoders}. As shown in Fig. \ref{FIG5}, the image and point cloud decoders share a similar design.  Our intuitions are to recover the spatial dimension using the pixel/point upsampling layers and decrease (or fuse) the feature channels using the ResNet/PointNet layers hierarchically.


For the image decoder shown in Fig. \ref{FIG5A}, we first concatenate ($\mathbf{F}_{\text{WP}}^{2}$, $\mathbf{F}_{\text{I}}^{2}$) and then feed them into a ResNet followed by a pixel upsampling operation (Res\&pixelUp) to obtain  feature map $\widetilde{\mathbf{F}}_{\text{I}}^{1}\in \mathbb{R}^{H_{1} \times W_{1} \times d_{1} '}$. Then we further concatenate ($\mathbf{F}_{\text{WP}}^{1}$, $\mathbf{F}_{\text{I}}^{1}$, $\widetilde{\mathbf{F}}_{\text{I}}^{1}$) and feed them into another two sets of such operators (Res\&pixelUp) to obtain the fused feature map $\widetilde{\mathbf{F}}_{\text{I}}\in \mathbb{R}^{H'\times W' \times d '}$. \ And $\widetilde{\mathbf{F}}_{\text{I}}$ will be passed into two CNNs for generating the pixel-wise scores $\mathbf{S}_{\text{I}} \in \mathbb{R}^{H'\times W'\times 1}$ 
and pixel-wise features $\mathbf{H}_{\text{I}} \in \mathbb{R}^{H'\times W'\times c}$. 
Finally, we determine that the pixels whose scores are higher than a threshold $\tau$ belong to the overlapping regions. Let $\mathbf{O}_{\text{I}}\in \mathbb{R}^{K_{\text{I}}\times 2}$ be the set of the coordinates of $K_{\text{I}}$ pixels detected in the overlapping regions, and $\mathbf{F}_{\mathbf{O}_{\text{I}}}\in \mathbb{R}^{K_{\text{I}}\times c}$ their features collected from $\mathbf{H}_{\text{I}}$.  



The point cloud decoder shown in Fig. \ref{FIG5B} shares the same procedure as the image decoder, except that ResNet is replaced with PointNet and the pixelUp with pointUp realized by PointNet++ \cite{POINTNET2}. 
The features are fed into a PointNet followed by a pointUp to generate the fused feature map $\widetilde{\mathbf{F}}_{\text{P}} \in \mathbb{R}^{N\times d '}$. Also, the CNNs are replaced with PointNet to generate the point-wise scores $\mathbf{S}_{\text{P}} \in \mathbb{R}^{N\times 1}$ and features $\mathbf{H}_{\text{P}} \in \mathbb{R}^{N\times c}$. We use the same threshold $\tau$ to filter them and obtain the estimated overlapping region $\mathbf{O}_{\text{P}}\in \mathbb{R}^{K_{\text{P}}\times 3}$ and the corresponding features $\mathbf{F}_{\mathbf{O}_{\text{P}}}\in \mathbb{R}^{K_{\text{P}}\times c}$, where $K_{\text{P}}$ is the number of overlapping points.


\subsection{Dense Correspondence-based Pose Estimation}
Under the assumption that the matched pixel-point pairs have similar features, whereas non-matched pairs have distinct features, we apply the nearest neighbor principle in the feature space of the detected overlapping region to establish 2D-3D correspondence.

Specifically, considering that multiple points cloud may be projected onto an identical pixel due to occlusions in the scene and the limited image resolution, we build the 2D-3D correspondence by finding the most similar pixel for each 3D point, i.e., 
for each point $\mathbf{P}_{i} \in \mathbf{O}_{\text{P}}$, $i=1,2,...,K_{\text{P}}$, we select the pixel in the $\mathbf{O}_{\text{I}}$ whose feature is the nearest to that of $\mathbf{P}_{i}$  as its correspondence. 
Formally, let $\mathbf{I}_{p(i)}$ 
be the corresponding pairs of pixels and points, where $p(i)$ is the index of the pixel in $\mathbf{O}_{\text{I}}$, obtained by optimizing 
\begin{equation}
p(i)=\mathop{\arg\min}_{j=1,2,...,K_{\text{I}}} \|\mathbf{F}_{\mathbf{O}_{\text{P}} ,i} -\mathbf{F}_{\mathbf{O}_{\text{I}} ,j} \|.
\end{equation}

It is inevitable that the above method will generate wrong correspondence, and directly applying the EPnP to them may decrease the registration accuracy, and even produce the wrong transformation. Similar to I2I and P2P registration using 2D or 3D correspondence, we run EPnP within RANSAC to optimize the camera pose and reject the outliers simultaneously. 

\subsection{Loss Function}
To drive the learning process of overlapping region detection and feature matching, we design a joint loss function consisting of a descriptor loss and a detector loss.  Generally, the descriptor loss promotes the network to produce similar features for matched pixel-point pairs and different features for non-matched pairs.
The detector loss encourages that the network can reliably identify overlap, producing higher scores for pixels and points inside the overlapping regions and lower scores for those outside the overlapping regions.\\

\noindent\textbf{Descriptor loss}. 
We use the cosine distance to compute the distance in the feature space, i.e., 
\begin{equation}
    d(\mathbf{F}_{1},\mathbf{F}_{2})=1-\frac{\left< \mathbf{F}_{1},\mathbf{F}_{2}\right>}{\|\mathbf{F}_{1}\|\|\mathbf{F}_{2}\|},
\end{equation}
where $\mathbf{F}_{1}$ and $\mathbf{F}_{2}$ represent image and point cloud feature vectors, and $\left<\cdot,\cdot\right>$ is their inner product.

Given  $\mathbf{I}$ and   $\mathbf{P}$, we use ground truth information to sample $n$ pairs of 2D-3D correspondence $\{\mathbf{I}_i,\mathbf{P}_i\}$, $i=1,2,...,n$ from the overlapping region of the image and point cloud, and their corresponding feature pairs are $\{\mathbf{F}_{\text{O}_{\text{I}},i},\mathbf{F}_{\text{O}_{\text{P}},i}\}$.
The feature distance of a positive pair is defined as
\begin{equation}
d_{\text{pos}}(i)=d(\mathbf{F}_{{\text{O}_{\text{I}}},i},\mathbf{F}_{{\text{O}_{\text{P}}},i}).
\end{equation}

If the distance between a pixel and the projection of a point onto the image is larger than a safe radius, denoted as $R$, they could be thought of as a negative pair. For a typical 3D point, many 2D pixels could be adopted to form a feasible negative pair.
Instead of using all these negative pairs, we only select the pair with the smallest feature distance and define the distance of a negative pair  as
\begin{equation}
d_{\text{neg}}(i)=\mathop{\min}_{j=1,2,...,n}\{d(\mathbf{F}_{{\text{O}_{\text{I}}},i},\mathbf{F}_{{\text{O}_{\text{P}}},j})\}\ s.t.\|\mathbf{I}_j-\mathbf{I}_i\|>R.
 \end{equation}

In order to accelerate the training process, we also introduce two margins, i.e., positive margin $M_\text{pos}$ and negative margin $M_\text{neg}$, and define the descriptor loss as 
\begin{equation}
\begin{split}
\mathcal{L}_{\text{desc}}=\frac{1}{n}\sum^{n}_{i=1}[\text{max}(0,d_{\text{pos}}(i)-M_{\text{pos}})+ \\ \text{max}(0,M_{\text{neg}}-d_{\text{neg}}(i))].
\end{split}
\end{equation}

\noindent\textbf{Detector loss}. 
According to the ground truth transformation, we sample $n$ pairs of pixels and points from the overlapping regions with their scores, $\{\mathbf{I}_u,\mathbf{S}_{\text{I},u}\}$, $\mathbf{I}_u\in \mathbf{O}_{\text{I}}$, $u=1,...,n$ and $\{\mathbf{P}_v,\mathbf{S}_{\text{P},v}\}$, $\mathbf{P}_v\in \mathbf{O}_{\text{P}}$, $v=1,...,n$. Besides, we also collect $n$ pixels and points in the non-overlapping region, $\{\mathbf{I}_k,\mathbf{S}_{\text{I},k}\}$, $\mathbf{I}_k\notin \mathbf{O}_{\text{I}}$, $k=1,..,n$ and $\{\mathbf{P}_m,\mathbf{S}_{\text{P},m}\}$, $\mathbf{P}_m\notin \mathbf{O}_{\text{P}}$, $m=1,...,n$. We define the detector loss as
 \begin{equation}
\mathcal{L}_{\text{det}}=\frac{1}{n}\left(-\sum_{u=1}^{n}\mathbf{S}_{\text{I},u}-\sum_{v=1}^{n}\mathbf{S}_{\text{P},v}+\sum_{k=1}^{n}\mathbf{S}_{\text{I},k}+\sum_{m=1}^{n}\mathbf{S}_{\text{P},m}\right).
 \end{equation}
Such a loss function promotes the pixels and points within the overlapping regions have high scores, while those beyond the overlapping regions have low scores during training.

In total, our loss function is
 \begin{equation}
\mathcal{L}=\mathcal{L}_{\text{desc}}+\lambda \mathcal{L}_{\text{det}},
 \end{equation}
where $\lambda$ is a hyperparameter for balancing the two terms.

\section{Experiments}
\label{sec:exp}

\subsection{Dataset}

We adopted two commonly used datasets, i.e., KITTI \cite{KITTI} and NuScenes \cite{NUSCENES}, to evaluate the proposed method.
\begin{itemize}
\item{\textbf{KITTI Odometry}\cite{KITTI}}.
The image-point cloud pairs were selected from the same data frame, i.e., the images and point clouds were captured simultaneously through a 2D camera and a 3D LiDAR with fixed relative positions. We followed the common practice \cite{DEEPI2P} of utilizing the first 8 sequences for training and the last 2 for testing. The transformation of the point cloud consists of a rotation around the up-axis and a 2D translation on the ground within the range of 10 m. During training, the image size was set to $160\times 512$, and the number of points to 40960.

\item{\textbf{NuScenes}\cite{NUSCENES}}.
Point clouds were acquired from a LiDAR with a 360-degree field of view. We used the official SDK to get the image-point cloud pairs, where the point cloud was accumulated from the nearby frames, and the image from the current data frame. We followed the official data split of NuScenes, where 850 scenes were used for training, and 150 for testing, and the transformation range was similar to the KITTI dataset. The image size was set to $160\times 320$, and the number of points was the same as the KITTI dataset.
\end{itemize}

\begin{table*}[h]
\centering
\renewcommand\arraystretch{1.5}
\setlength{\tabcolsep}{4mm}{
\caption{Comparison of the registration accuracy (mean $\pm$ std) of different methods on the KITTI and NuScenes datasets. ``$\downarrow$" means that the smaller, the better. The best results are highlighted in bold.}
\label{TAB1}
\begin{tabular}{l|c|c|c|c}
\hline
\hline
\multirow{2}{0.1\textwidth}{~} & \multicolumn{2}{c|}{KITTI} &  \multicolumn{2}{c}{NuScenes} \\
\cline{2-5}
  & RTE $\downarrow$ (m) & RRE $\downarrow$ ($^{\circ}$) & RTE $\downarrow$ (m) & RRE $\downarrow$ ($^{\circ}$) \\
\hline
Grid Cls. + EPnP \cite{DEEPI2P} & $1.07\pm 0.61$ & $6.48\pm1.66$ & $2.35\pm1.12$ & $7.20\pm1.65$ \\
\hline
DeepI2P (3D) \cite{DEEPI2P} & $1.27\pm0.80$ & $6.26\pm2.29$ & $2.00\pm1.08$ & $7.18\pm1.92$ \\
DeepI2P (2D) \cite{DEEPI2P} & $1.46\pm0.96$ & $4.27\pm2.74$ & $2.19\pm1.16$ & $3.54\pm2.51$ \\
\hline
Ours & $\pmb{0.74\pm0.65}$ & $\pmb{2.07\pm1.64}$ & $\pmb{1.83\pm1.06}$ & $\pmb{2.65\pm1.93}$ \\
\hline
\hline
\end{tabular}}
\end{table*}

\subsection{Implementation Details}
\noindent\textbf{Training}.
During training, we established the ground truth 2D-3D correspondence to supervise the network. The transformation of a 3D point $\mathbf{P}_i\in \mathbb{R}^3$  to the image coordinate of the camera $\mathbf{p}_{i}\in \mathbb{R}^2$ is given by
\begin{equation}
    \tilde{\mathbf{p}}_{i}=\begin{bmatrix}x_i' \\ y_i' \\ z_i'\end{bmatrix}=\mathbf{K}(\mathbf{R}_\text{gt}\mathbf{P}_{i}+\mathbf{t}_\text{gt}),
 \end{equation}
 \begin{equation}
 \mathbf{p}_{i}=\begin{bmatrix}p_{xi}\\p_{yi} \end{bmatrix}=\begin{bmatrix}x_i'/z_i'\\y_i'/z_i' \end{bmatrix},
 \end{equation}
where $\mathbf{K}\in\mathbb{R}^{3\times3}$ is the camera intrinsic matrix and $\mathbf{T}_\text{gt}=[\mathbf{R}_\text{gt}|\mathbf{t}_\text{gt}]$ is the ground truth camera pose.

For image features, $\mathbf{F}^{1}_\text{I}$ is a 1/16 scale of the original image while $\mathbf{F}^{2}_\text{I}$ is 1/32, and $\mathbf{H}_{\text{I}}$ is 1/4, i.e., $W=16W_1=32W_2=4W'$ and $H=16H_1=32H_2=4H'$, as shown in Fig. \ref{FIG3}. As for point cloud features, $N_1=N_2=256$, and the number of the k-NN in SO-Net \cite{SONET} is $k=32$. The feature channels are $d_1=d_1'=256$, $d_2=d_3=512$ and $c=128$. We trained our network for 25 epochs on each dataset, with a batch size of 24. We used the Adam \cite{ADAM} to optimize the network, and the initial learning rate was 0.001 and multiplied by 0.25 every five epochs until it reached 0.00001. During training, we set the safe radius $R$ to 1 pixel, the value of $\lambda$ involved in the loss function to 0.5, the positive margin to $M_{pos}=0.2$, and the negative margin to $M_{neg}=1.8$. 
\\

\noindent\textbf{Testing}.
Based on the experimental observation, we set $\tau=0.9$ to determine the overlapping regions. 
For RANSAC, we set the number of iterations to
500 and the threshold for inlier reprojection error to 1 pixel.

\subsection{Compared Methods}
We compared our CorrI2P with  the setting called Grid Cls. + PnP. and Frus. Cls. + Inv.Proj.:
\begin{itemize}
\item{\textbf{Grid Cls. + PnP.}}  proposed in recent DeepI2P \cite{DEEPI2P} divides the image into 32$\times$32 grids and uses the network to classify which grid the points in the point cloud would be projected into.
It establishes 2D-3D correspondence based on the classification result and then uses EPnP with RANSAC to estimate the camera pose.

\item{\textbf{Frus. Cls. + Inv.Proj.}}  uses the frustum classification, and inverse camera projection in DeepI2P \cite{DEEPI2P} to obtain the camera pose. We used the same network setting as their paper and tried the 2D and 3D inverse camera projection to optimize the pose, namely DeepI2P (2D) and DeepI2P (3D), respectively. We used a 60-fold random initialization to search for the initial pose of the camera. \\

\end{itemize} 
\textbf{Evaluation metric}.
Similar to P2P registration \cite{METRIC}, we used Relative Translational Error (RTE) $E_\text{t}$ and Relative Rotational Error (RRE) $E_\text{R}$  to evaluate our registration result, respectively computed as 
\begin{equation}
E_\text{R}=\sum_{i=1}^{3}|\gamma(i)|, 
\end{equation}
\begin{equation}
E_\text{t}=\|\mathbf{t}_{\text{gt}}-\mathbf{t}_{\text{E}}\|,
\end{equation}
where $\gamma$ is the Euler angle of the matrix $\mathbf{R}_{\text{gt}}^{-1}\mathbf{R}_{\text{E}}$, $\mathbf{R}_{\text{gt}}$ and $\mathbf{t}_{\text{gt}}$ are the ground-truth transformation, and $\mathbf{R}_{\text{E}}$ and $\mathbf{t}_{\text{E}}$ are the estimated transformation.

In the ablation study, we also conducted a feature matching experiment to show the quality of the correspondence estimator. Inspired by  P2P registration \cite{RECALL}, we designed two kinds of recall to evaluate the feature matching. Pair recall $R_\text{pair}$ is the ratio of the correct correspondences, while fragment recall $R_\text{frag}$ is the ratio of the fragments with higher proportion of correct 2D-3D correspondences than a pre-set threshold. They are calculated as
\begin{equation}
\begin{aligned}
    R_\text{pair}=\frac{1}{M}\sum_{s=1}^{M}\left(\frac{1}{|\mathbf{O}_\text{P}^s|}\sum_{i\in \mathbf{O}_\text{P}^s}\mathbbm{1}\left(\|\mathbf{p}_{p(i)}^s- 
    \pi(\mathbf{T}_\text{gt}^s\mathbf{P}_i^s)\|<\tau_1\right) \right ),
\end{aligned}
\end{equation}

\begin{equation}
\begin{aligned}
     R_\text{frag}=\frac{1}{M}\sum_{s=1}^{M}\mathbbm{1}\left(\left(\frac{1}{|\mathbf{O}_\text{P}^s|}\sum_{i\in \mathbf{O}_\text{P}^s}\mathbbm{1}\left(\|\mathbf{p}_{p(i)}^s-  \right. \right. \right.
     \\   \left.  \pi(\mathbf{T}_\text{gt}^s\mathbf{P}_i^s)\|<\tau_1 \Big)\Bigg)>\tau_2\right),
\end{aligned}
\end{equation}
where $\tau_1$ and $\tau_2$ are the inlier distance and inlier ratio threshold, $M$ is the number of the ground truth matching image-point cloud pairs, $\pi(\cdot)$ is the projection from the point cloud to image coordinate according to Eqs. (8)-(9), $\mathbbm{1}(\cdot)$ is the indicator function, $\mathbf{T_\text{gt}^s}$ is the ground truth transformation of the $s$-th image-point cloud pair, $\mathbf{p}$ is the pixel coordinate, and $\mathbf{P}$ is the point cloud coordinate. $p(i)$ is the index of $i$-th point in the point cloud's corresponding pixel using Eq. (1).

\begin{figure*}[h]
\centering
\subfloat[KITTI dataset]{\label{FIG6A}
\includegraphics[width=0.5\textwidth]{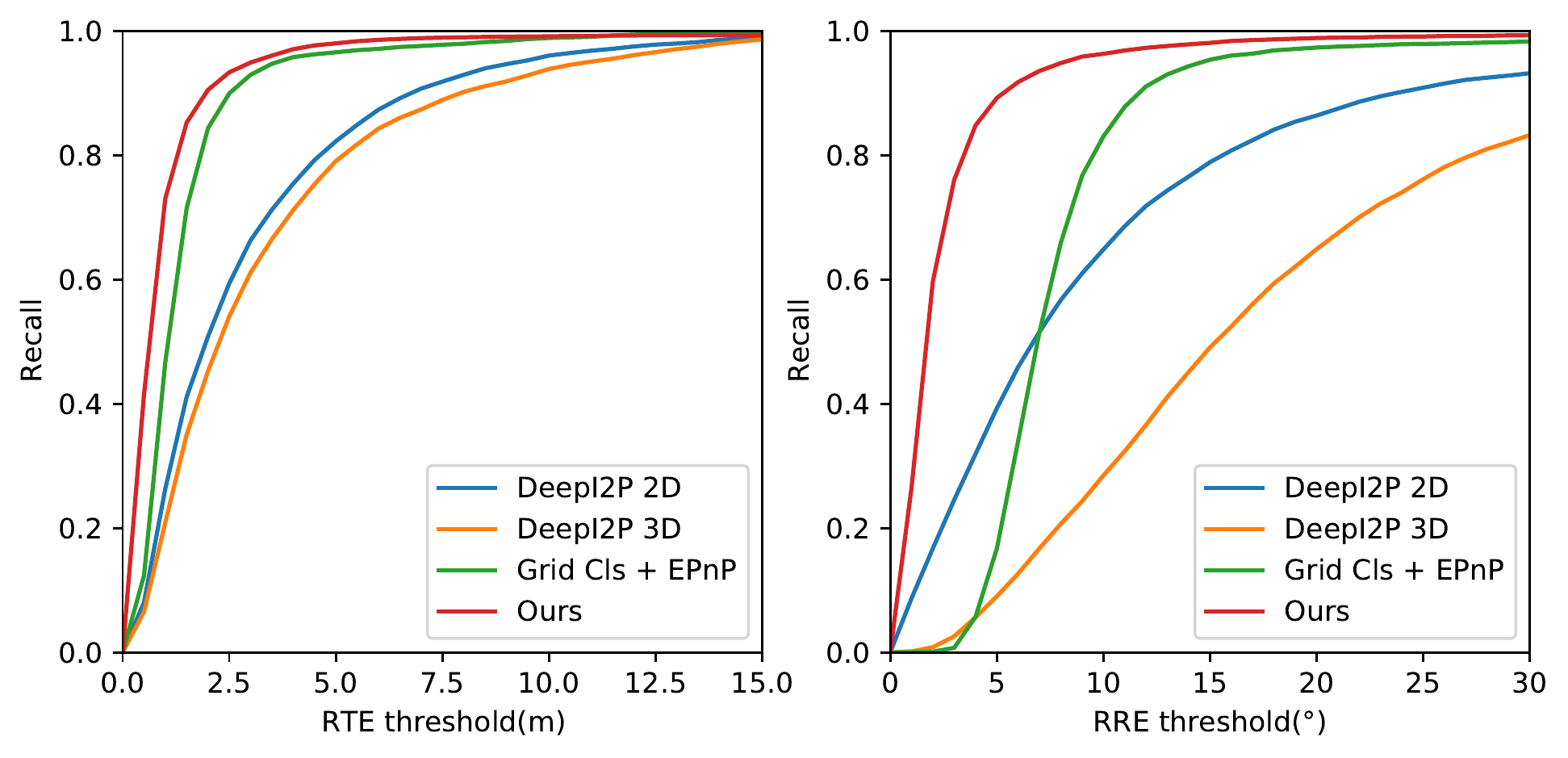}}
\subfloat[NuScenes dataset]{\label{FIG6B}
\includegraphics[width=0.5\textwidth]{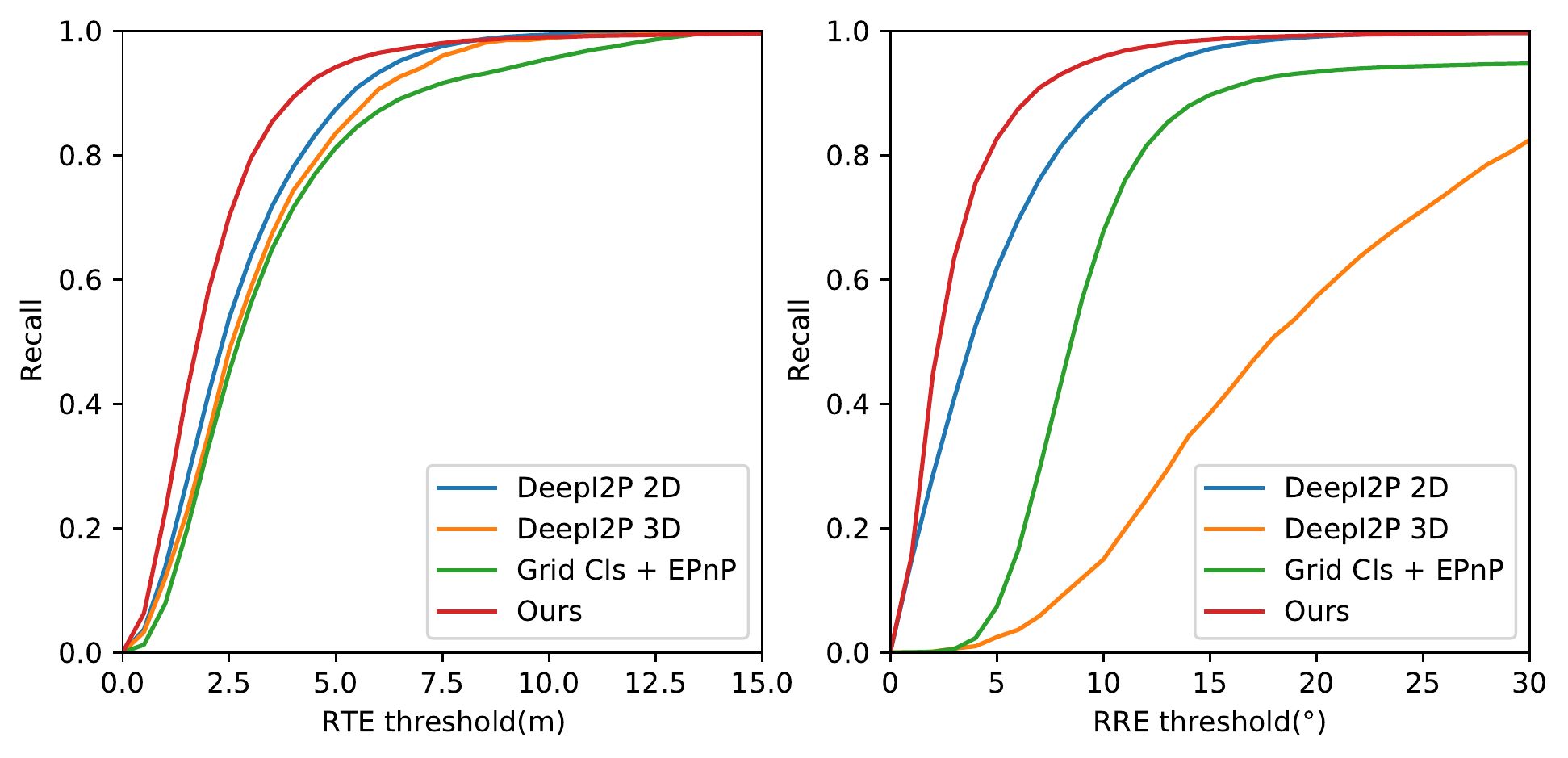}}
\caption{Comparison of the Registration recall of different methods with various RTE and RRE thresholds on KITTI  and NuScenes datasets.}
\label{FIG6}
\end{figure*}

\begin{figure*}[!htbp]
\centering 

{%
\begin{tikzpicture}[]

\node[] (a) at (0,3.2) {\includegraphics[width=0.16\textwidth]{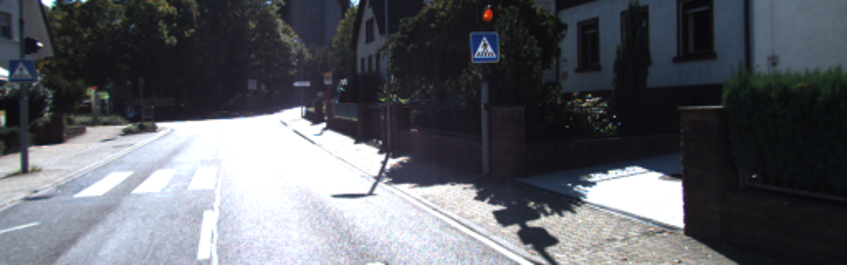} };
\node[] (a) at (0,2.5) { (a) };
\node[] (f) at (0,1.2) {\includegraphics[width=0.16\textwidth]{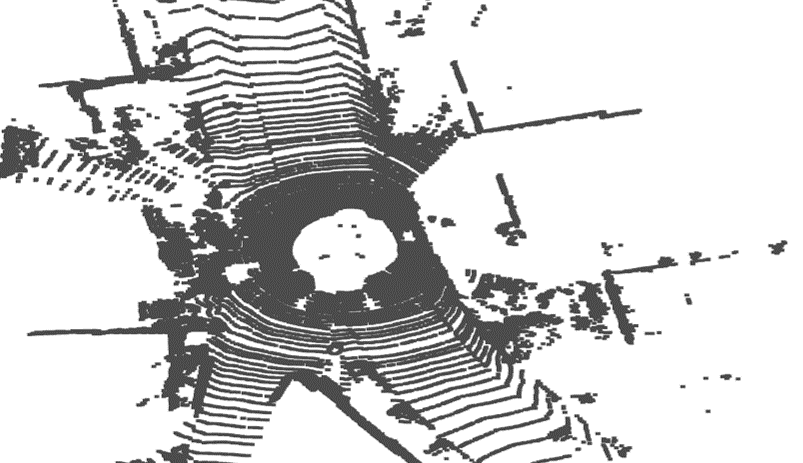}};
\node[] (f) at (0,0) { (f) };

\node[] (b) at (3.05,3.2) {\includegraphics[width=0.16\textwidth]{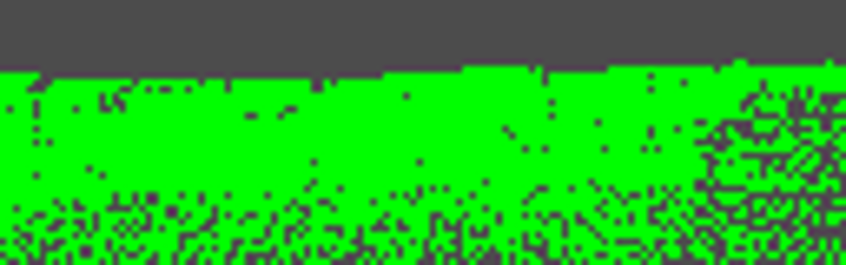}};
\node[] (b) at (3.05,2.5) { (b) };
\node[] (g) at (3.05,1.2) {\includegraphics[width=0.16\textwidth]{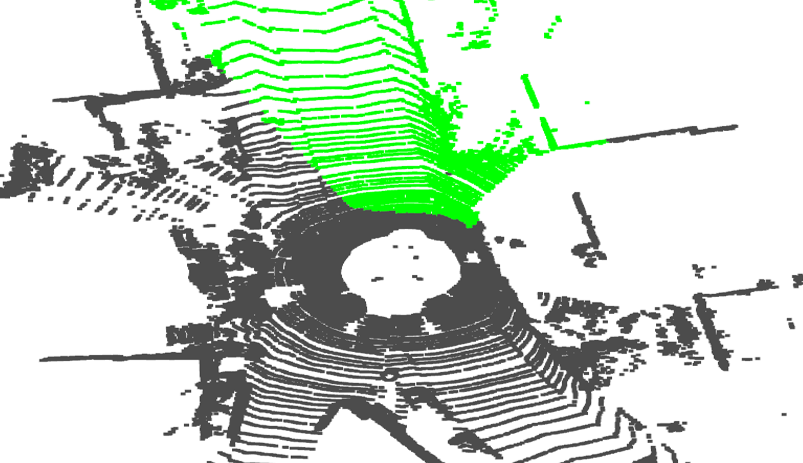}};
\node[] (g) at (3.05,0) { (g) };

\node[] (c) at (6.1,3.2) {\includegraphics[width=0.16\textwidth]{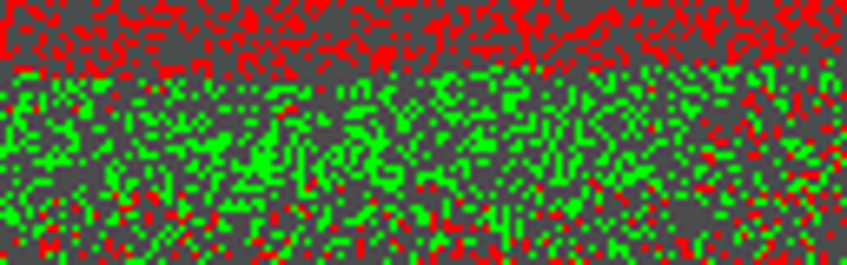}};
\node[] (c) at (6.1,2.5) { (c) };
\node[] (h) at (6.1,1.2) {\includegraphics[width=0.16\textwidth]{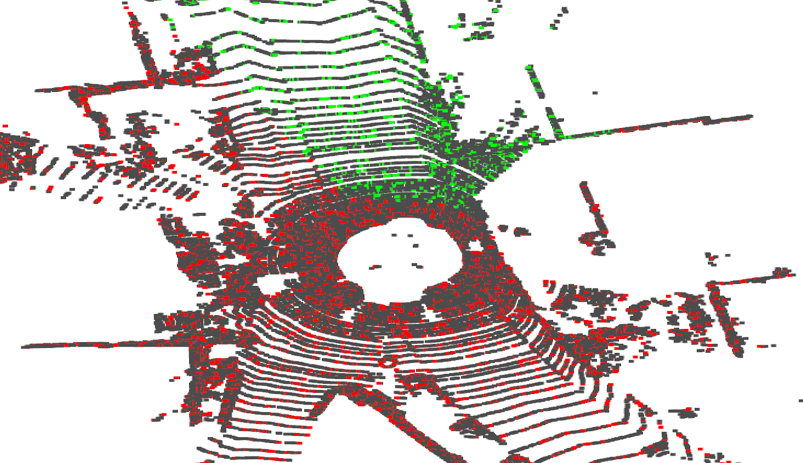}};
\node[] (h) at (6.1,0) { (h) };

\node[] (d) at (9.15,3.2) {\includegraphics[width=0.16\textwidth]{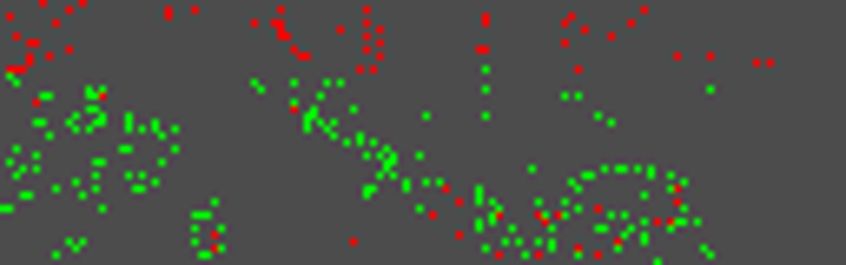}};
\node[] (d) at (9.15,2.5) { (d) };
\node[] (i) at (9.15,1.2) {\includegraphics[width=0.16\textwidth]{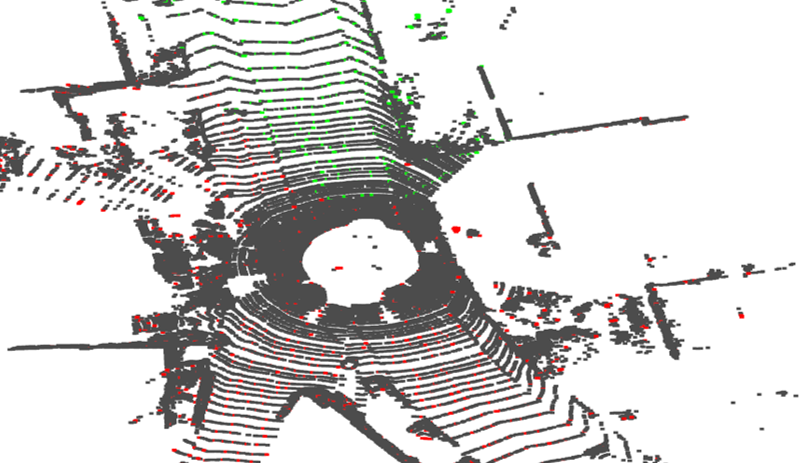}};
\node[] (i) at (9.15,0) { (i) };

\node[] (e) at (12.20,3.2) {\includegraphics[width=0.16\textwidth]{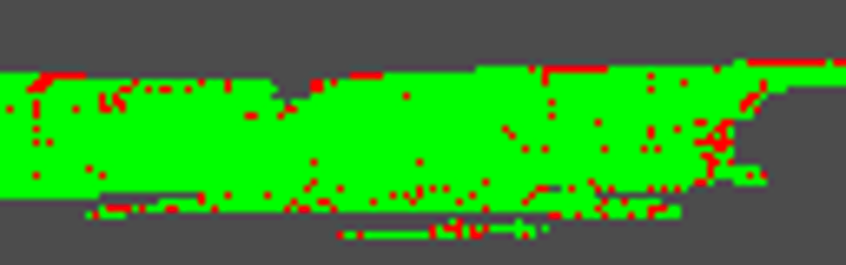}};
\node[] (e) at (12.20,2.5) { (e) };
\node[] (j) at (12.20,1.2) {\includegraphics[width=0.16\textwidth]{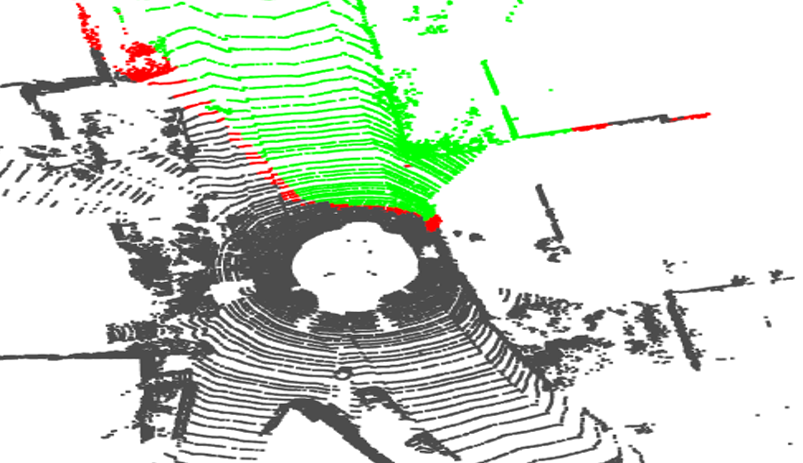}};
\node[] (j) at (12.20,0) { (j) };
\node[] (k) at (15.25,1.8) {\includegraphics[width=0.16\textwidth]{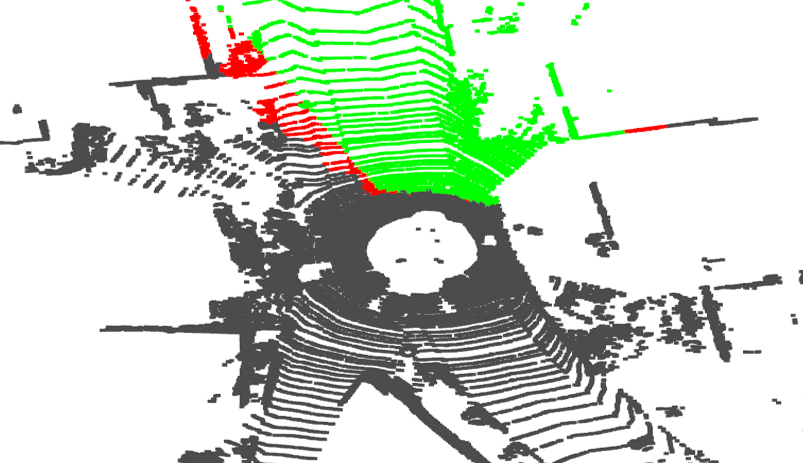}};
\node[] (k) at (15.25,0.6) { (k) };
\end{tikzpicture}
}
\setlength{\abovecaptionskip}{-0.55cm}
\caption{Visual illustration of overlapping region detection on the image and point cloud. The green pixels and points are detected in the overlapping region of the image and point cloud, and the red ones are wrongly detected. Note that DeepI2P conducts overlapping region detection only on point clouds.
It is obvious that our method has better performance on both image and point cloud overlapping region detection over other methods. (a) and (f): Original. (b) and (g): Ground Truth. (c) and (h): Random. (d) and (i): SIFT-ISS \cite{2D3DMATCHNET}. (e) and (j): Ours. (k): DeepI2P \cite{DEEPI2P}.} 
\label{overlap}
\label{overlap:ori:img}
\label{overlap:ori:pc}
\label{overlap:gt:img}
\label{overlap:gt:pc}
\label{overlap:deepi2p:pc}
\label{overlap:ours:pc}
\label{overlap:ours:img}
\label{overlap:iss:pc}
\label{overlap:sift:img}
\label{overlap:random:pc}
\label{overlap:random:img}

\end{figure*}
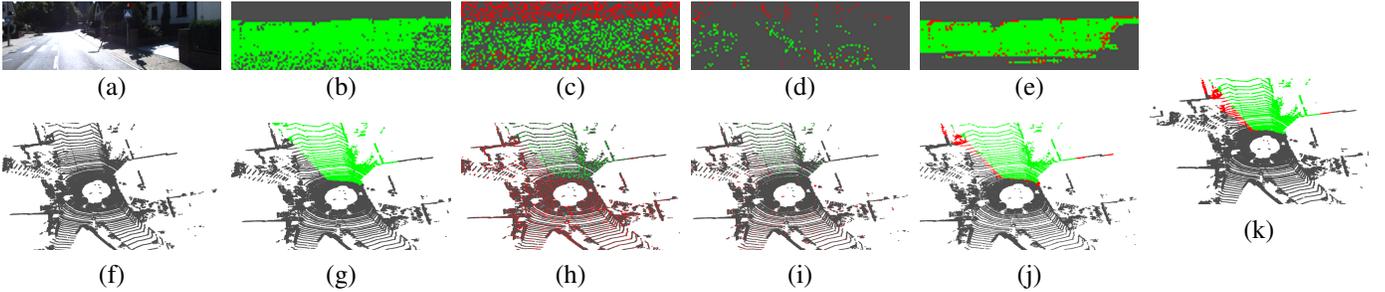

\begin{figure*}[!htbp]
\centering
\subfloat[RTE(m)/RRE($^{\circ}$)]{\label{corr:no}
\includegraphics[width=0.193\textwidth]{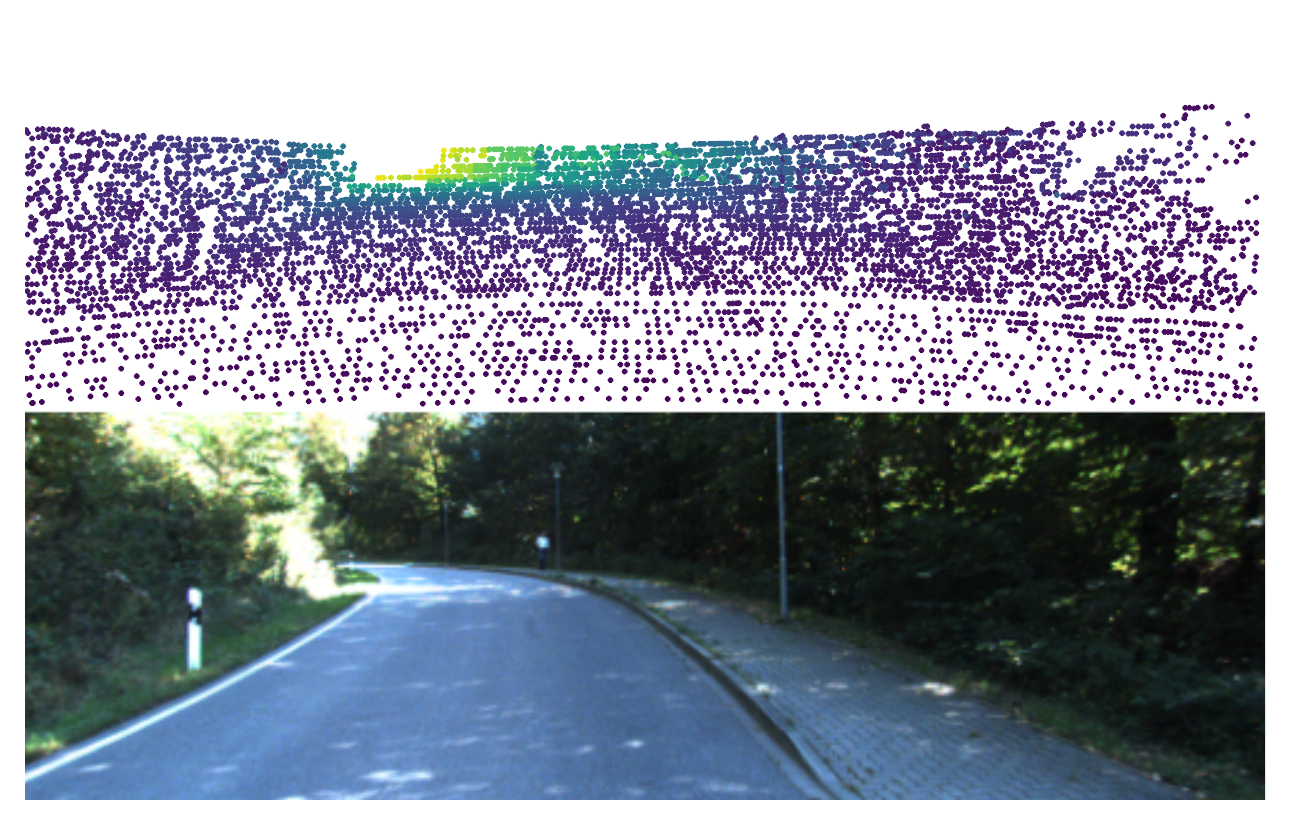}}
\subfloat[2.90/9.16]{\label{corr:random}
\includegraphics[width=0.193\textwidth]{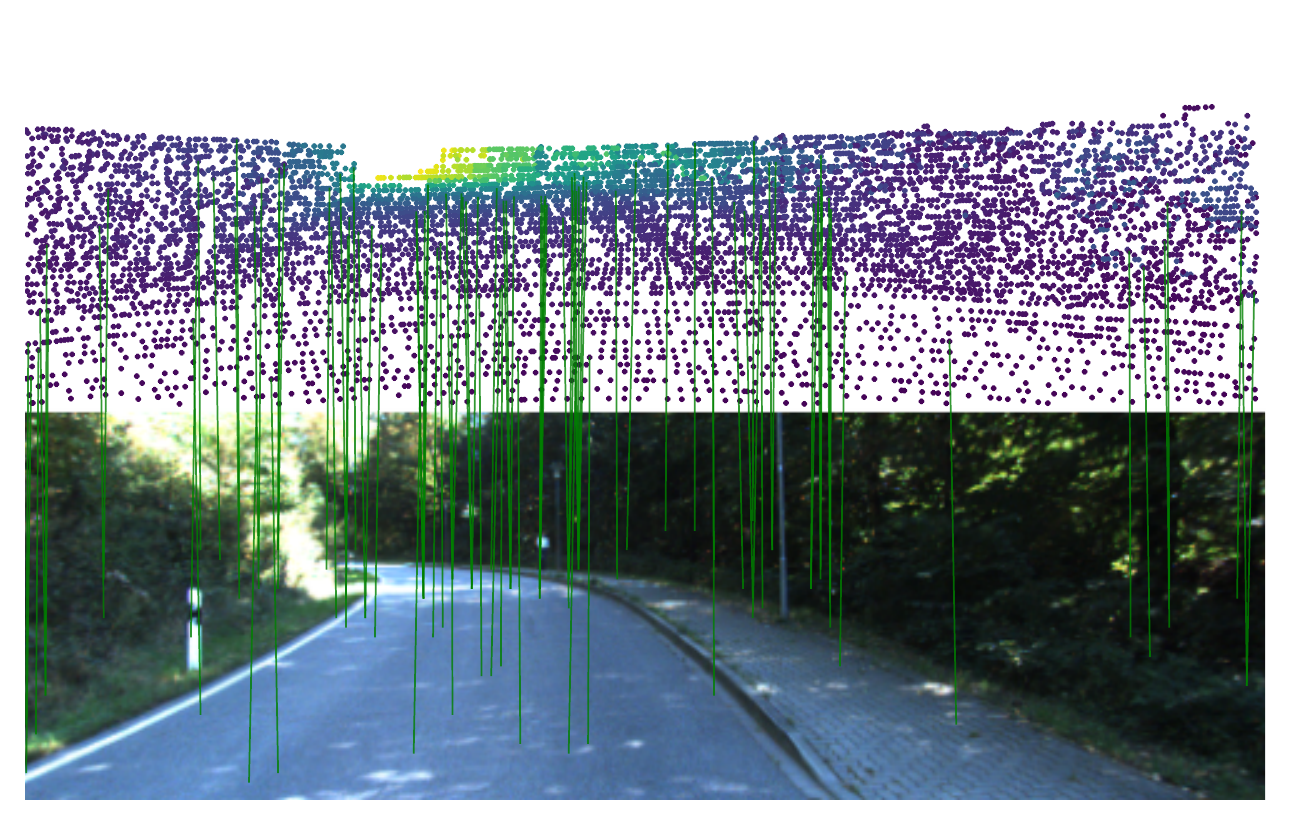}}
\subfloat[2.76/5.73]{\label{corr:sift:iss}
\includegraphics[width=0.193\textwidth]{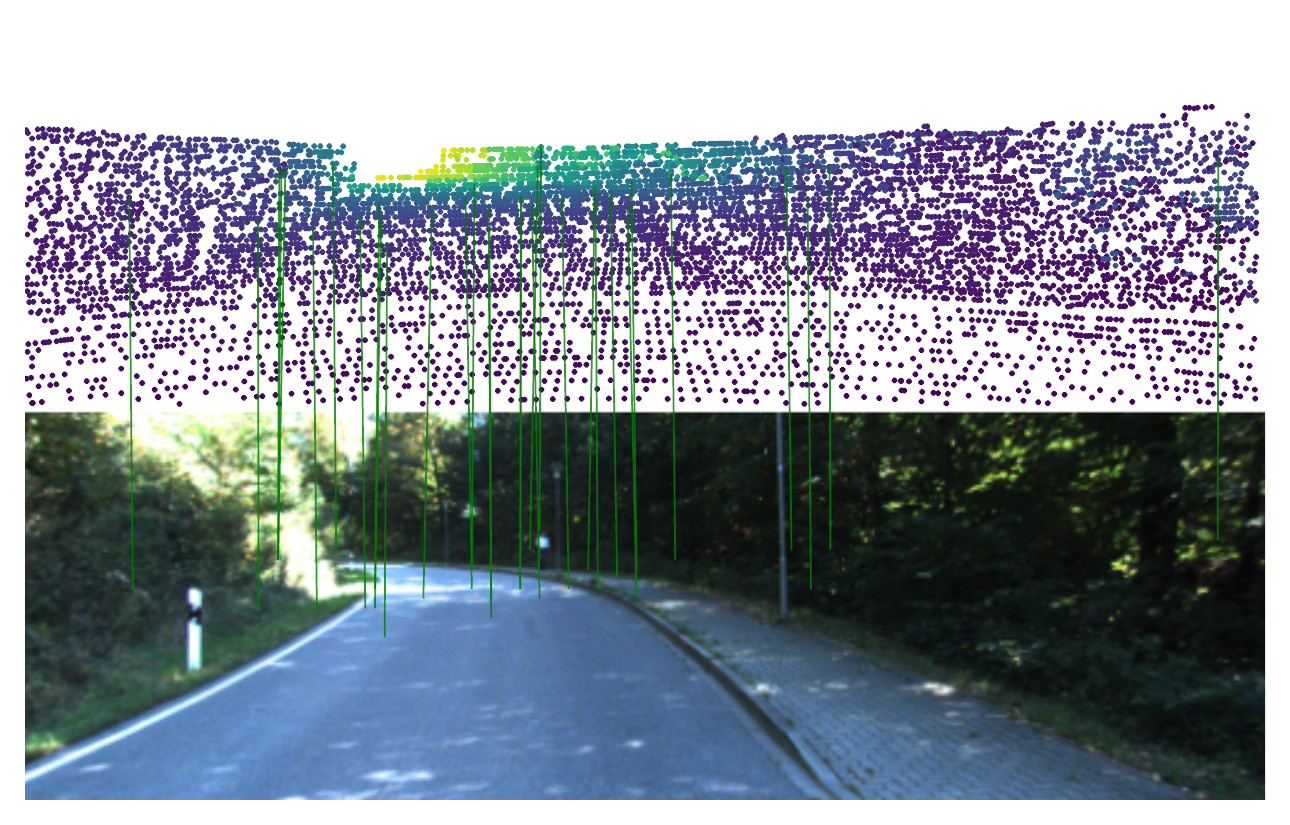}}
\subfloat[1.63/8.02]{\label{corr:grid:cls}
\includegraphics[width=0.193\textwidth]{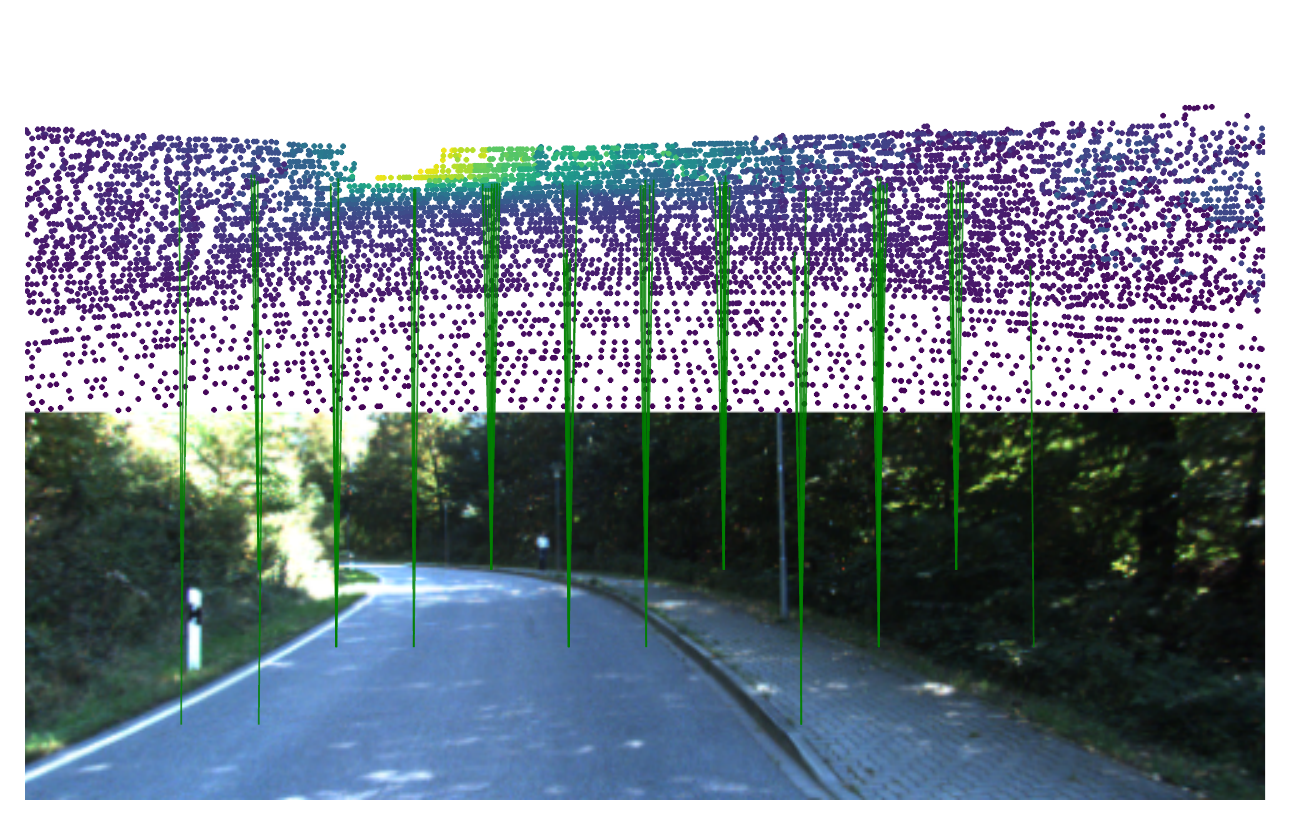}}
\subfloat[0.66/0.49]{\label{corr:ours}
\includegraphics[width=0.193\textwidth]{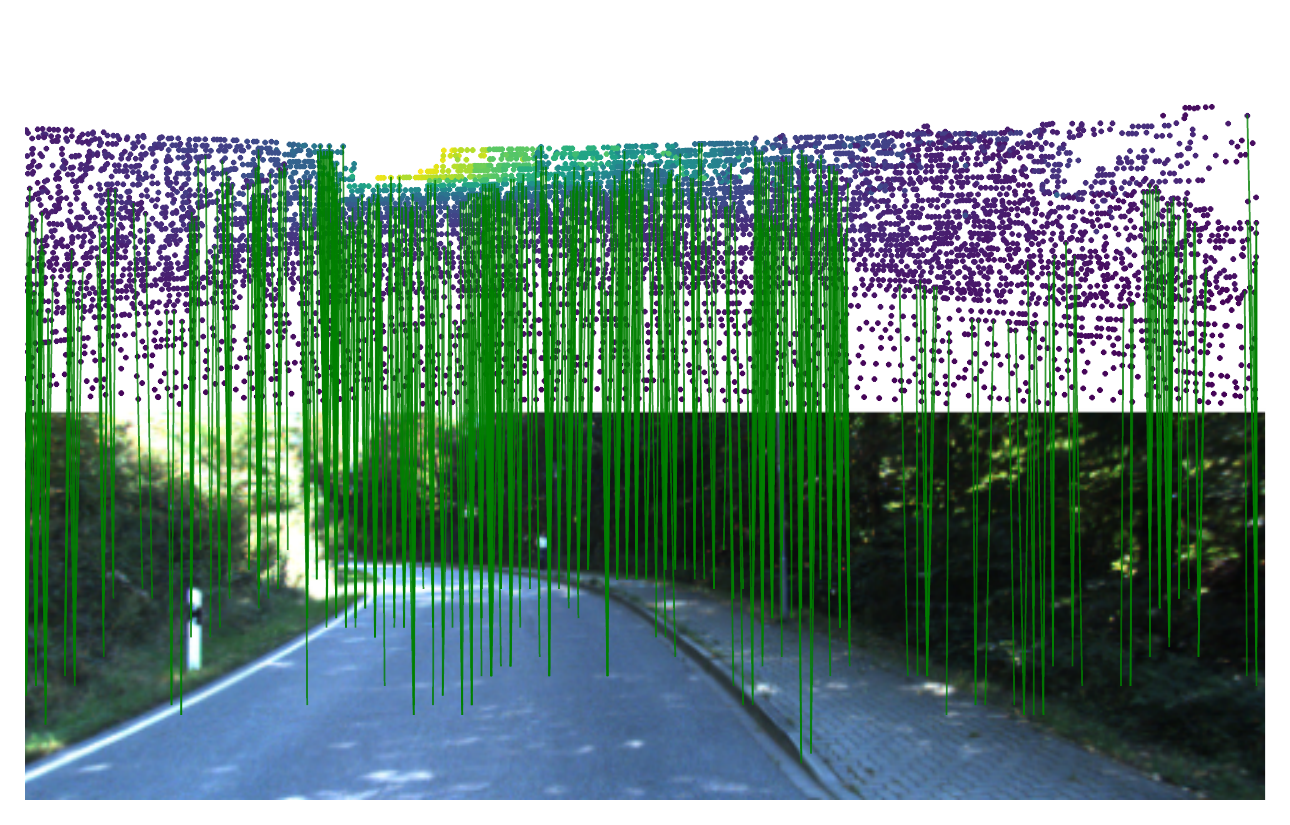}}
\caption{Comparison of 2D-3D correspondence and registration accuracy  
by different methods. We aligned the point cloud 
through ground-truth transformation for visualization purposes, the color of points represents the depth, and the lines represent the 2D-3D correspondence. Here we only show the correct correspondence. It can be seen that our method produces the densest correspondence and the highest registration accuracy. (a) Original. (b) Random. (c) SIFT-ISS \cite{2D3DMATCHNET}. (d) Grid Cls \cite{DEEPI2P}. (e) Our method.}
\label{corr}
\end{figure*}

\subsection{Results}

\begin{table}[t]
    \centering
    \renewcommand\arraystretch{1.5}
    \caption{The performance of the overlap detection on the KITTI dataset. The best results are highlighted in bold.}
    \begin{tabular}{c|l|c|c|c}
    \hline\hline
    ~ & Method & Recall & Precision & F2-Score \\
    \hline
      \multirow {4} {*}  {PC}  & ISS \cite{ISS} & 0.044 & 0.268 & 0.076\\
      & Random & 0.199 & 0.196 & 0.197\\
      & DeepI2P \cite{DEEPI2P} & 0.935 & 0.946 & 0.938 \\
      & Ours  & 0.975 & 0.911 & \pmb{0.941} \\
    \hline
     \multirow {3} {*}  {IMG} & SIFT \cite{SIFT} & 0.091 & 0.585 & 0.156 \\
        & Random  & 0.329 & 0.599 & 0.424 \\
        & Ours  & 0.783 & 0.903 & \pmb{0.838} \\
    \hline\hline
    \end{tabular}
\label{overlapdetection}
\end{table}

\noindent\textbf{Registration accuracy}.  Since some failed registration results may cause dramatically large RRE and RTE, showing unreliable error metrics, similar to P2P registration \cite{POINTDSC,HREGNET}, we calculated the average RTE and RRE only for those with RTE lower than 5m and RRE lower than 10$^{\circ}$. The registration accuracy is illustrated in Table \ref{TAB1}, where it can be seen that our method outperforms all compared methods by a noticeable margin on both datasets. Besides, for a more detailed comparison of the registration performance, we showed the registration recall with different RTE and RRE thresholds on two datasets in Fig. \ref{FIG6}.

As listed in Table \ref{overlapdetection}, although the recall, precision, and F2-Score values of the frustum classification of DeepI2P \cite{DEEPI2P} achieve 0.935, 0.946, and 0.938, respectively,  it is still worse than our CorrI2P because the points located in the boundary of the frustum are prone to be wrongly classified, as shown in Fig. \ref{overlap:deepi2p:pc}, which has an adverse influence on the inverse camera projection and eventually gets the wrong camera pose. Grid Cls. + EPnP has worse registration accuracy because the $32\times 32$ grid size is too coarse to get an accurate pose, and in this way, it can only get sparse and coarse 2D-3D correspondence, as shown in Fig. \ref{corr:grid:cls}, which decreases the registration accuracy although the grid classification accuracy is higher than 0.50. By contrast, our CorrI2P estimates the camera pose according to dense correspondence, as shown in Fig. \ref{corr:ours}, which is beneficial to the final registration accuracy. \\

\noindent\textbf{Error distribution}.
The distributions of the registration error, i.e., RTE and RRE, are shown in Fig. \ref{FIG8}, where it can be seen that the performance is better on KITTI than NuScenes. The mode of RTE/RRE is $\sim$0.5m/2° on KITTI and $\sim$1.5m/2° on NuScenes. The RTE and RRE variances are also smaller on KITTI. \\

\noindent\textbf{Accuracy of overlapping region detection}. Overlapping region detection is critical for our method to select the corresponding pixels and points, and the accurate overlapping region detection would increase the registration accuracy.
As visualized in Fig. \ref{overlap}, the overlapping region predicted by our method is the most accurate. Furthermore, we conducted experiments to quantitatively compare the accuracy of overlapping region detection on the KITTI dataset. As overlapping region detection on the image and point cloud can be regarded as pixel-wise and point-wise binary classification,
we used recall, precision, and F2-score as metrics to evaluate the performance of our overlapping region detection. We adopted random sampling on the image and point cloud as a baseline,
where 2048 pixels and 8192 points were sampled from the image and point cloud, respectively. Besides, we also used SIFT \cite{SIFT} and ISS \cite{ISS} to extract the keypoints in the image and point cloud and regarded them as overlapping regions, just like 2D3D-MatchNet \cite{2D3DMATCHNET}. DeepI2P uses  point-wise classification to select the points within the frustum, i.e., the overlapping region of the point cloud, so we used it as a comparison of the overlapping region detection on the point cloud.

The results are listed in Table \ref{overlapdetection}, where it can be seen that the detection accuracy of Random, SIFT, and ISS is much worse than ours. The reason is that these methods select pixels and points in the whole image and point cloud, but their ground truth overlapping regions only take up a small proportion, resulting in less correct correspondence and lower registration accuracy. The ablation study would show the accuracy of registration based on these overlapping region detection methods. And it can also be seen that our overlapping region detection method for the point cloud is better than DeepI2P. The precision of our overlapping region detection is higher than 0.9 on both image and point cloud, ensuring registration accuracy. \\

\begin{figure*}[h]
\centering
\subfloat[Feature Matching Recall]{\label{FIG7A}
\includegraphics[width=0.5\textwidth]{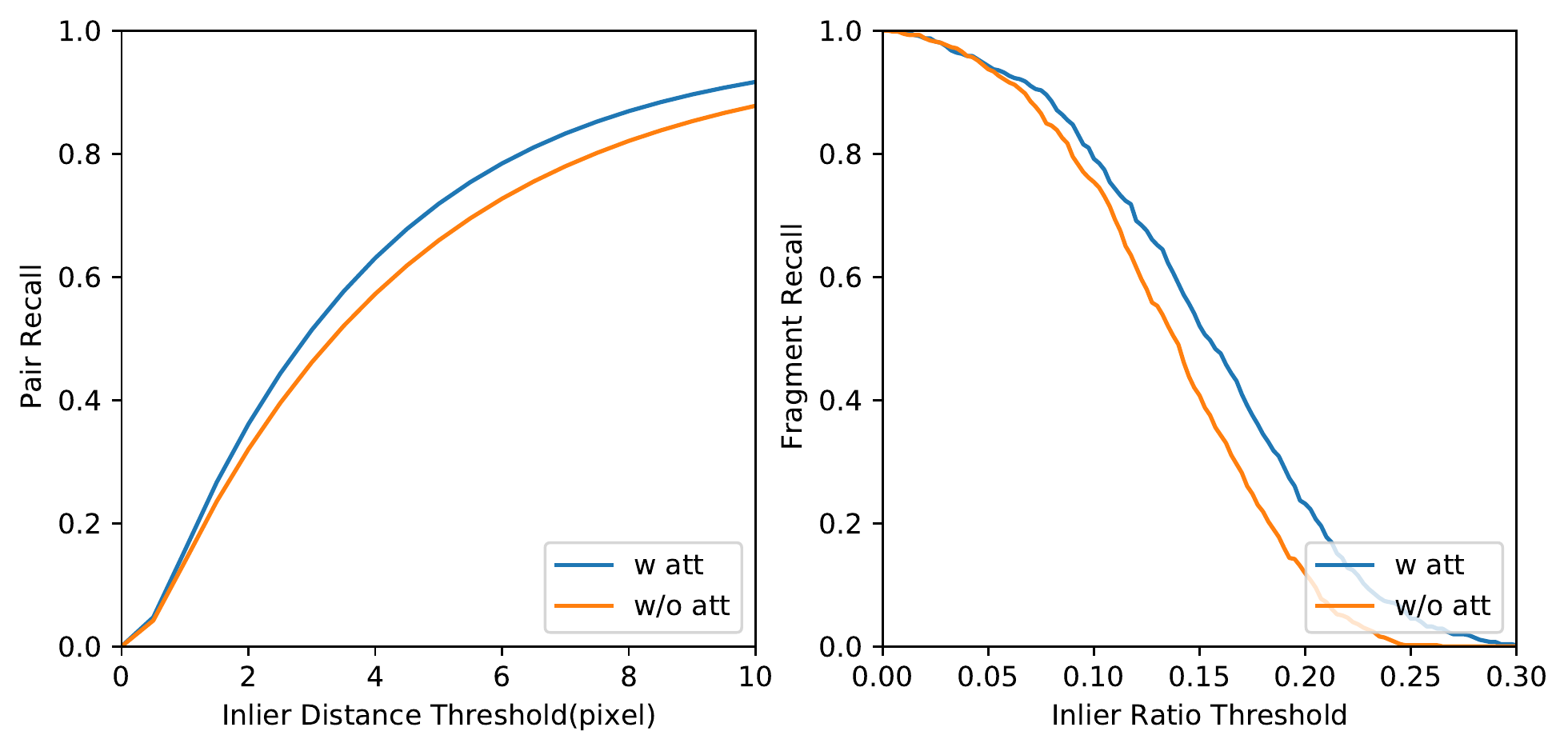}}
\subfloat[Registration Recall]{\label{FIG7B}
\includegraphics[width=0.5\textwidth]{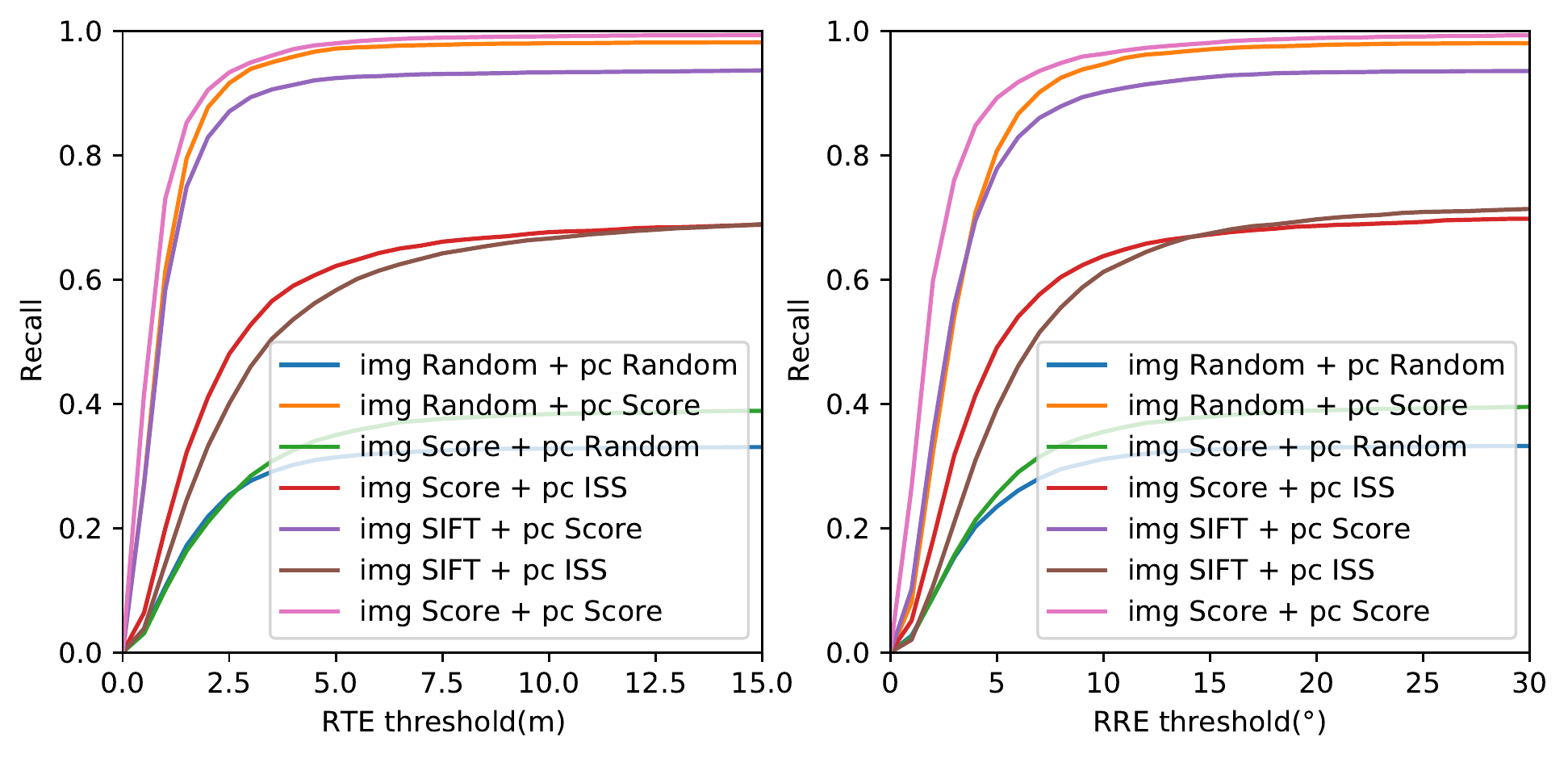}}
\caption{Ablation study results on the KITTI dataset. (a) Feature matching recall in relation to inlier ratio threshold $\tau_1$(left) and inlier ratio threshold $\tau_2$(right). (b) Registration recall with different RTE and RRE thresholds.}
\label{FIG7}
\end{figure*}

\begin{figure}[h]
\centering
\includegraphics[width=0.5\textwidth]{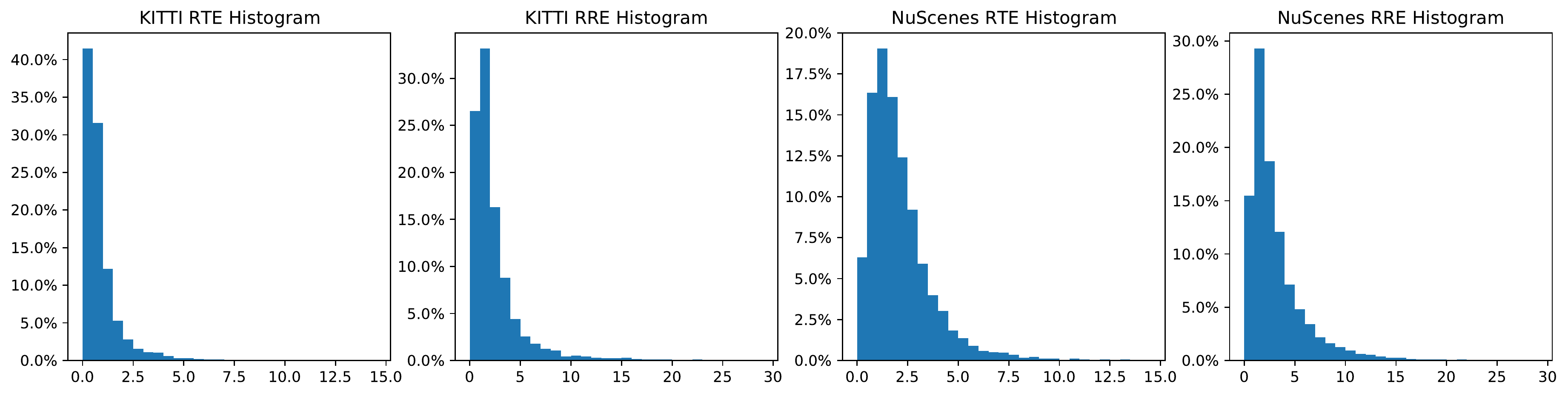}
\caption{Histograms of RTE and RRE on the KITTI and NuScenes datasets. x-axis is RTE(m) and RRE(°), and y-axis is the percentage.}
\label{FIG8}
\end{figure}

\begin{table}[t]
    \centering
    \renewcommand\arraystretch{1.5}
    \caption{Comparison of the registration accuracy of different sampling strategies on the KITTI dataset. The best results are highlighted in bold.}
    \begin{tabular}{c|c|c|c}
    \hline\hline
      IMG & PC & RTE (m) & RRE ($^{\circ}$) \\
    \hline
    Random & Random   & $1.57\pm 1.01$ & $3.53\pm 2.22$ \\
    \hline
    Random & Score  & $0.94\pm 0.71$ & $3.06\pm 1.85$ \\
    \hline
    Score & Random & $1.79\pm 1.14$ & $3.67\pm 2.19$ \\
    \hline
    Score & ISS  & $1.67\pm1.08$ & $3.40\pm2.12$  \\
    \hline
    SIFT & Score & $0.94\pm0.71$ & $2.89\pm1.88$\\
    \hline
    SIFT & ISS &$1.91\pm1.15$ &$4.13\pm2.25$\\
    \hline
    Score & Score& $\pmb{0.74\pm0.65}$ & $\pmb{2.07\pm1.64}$ \\
    \hline\hline
    \end{tabular} 
\label{TAB2}
\end{table}

\begin{table}[t]
    \centering
    \renewcommand\arraystretch{1.5}
    \caption{Comparison of the registration accuracy of different 3D point density.}
    \begin{tabular}{c|c|c}
    \hline\hline
      \# points & RTE (m) & RRE ($^{\circ}$) \\
    \hline
    5120    & $1.19\pm 0.87$ & $2.72\pm 1.91$ \\
    \hline
    10240   & $1.08\pm 0.83$ & $2.59\pm 1.91$ \\
    \hline
    20480  & $0.93\pm 0.77$ & $2.26\pm 1.77$ \\
    \hline
    40960   & $0.74\pm0.65$ & $2.07\pm1.64$  \\
    \hline\hline
    \end{tabular} 
\label{TAB3}
\end{table}


\noindent\textbf{KITTI vs. NuScenes}.
Our registration accuracy is higher on   KITTI than that on NuScenes. The main reason is that the point clouds of the scene in the two datasets are acquired through different methods. The point cloud in the KITTI dataset is dense enough to use directly.
As for the NuScenes dataset, every point cloud frame is so sparse that we need to splice it with the adjacent frames. However, the point cloud is collected on the street, and some components in the scenes are dynamic, such as cars or pedestrians, making the point cloud not aligned completely, which would cause trouble extracting point cloud features.

\begin{table*}[h]
    \centering
    \renewcommand\arraystretch{1.5}
    \caption{Comparison of the efficiency of different methods on the KITTI dataset.}
    \begin{tabular}{l|c|c|c|c|c}
    \hline\hline
     Method &  Network size (MB) & FLOPs (G) & GPU Memory (GB) &Inference (ms) & Pose Estimation (s)  \\
    \hline
    Grid Cls. + EPnP   &  100.75  & 20.75  & 2.39  & $11.20$   & $0.04$\\
    \hline
    DeepI2P (3D)  & 100.12 & 13.99 & 2.01  &$7.55$    & $16.58$\\
    DeepI2P (2D) & 100.12 & 13.99 & 2.01  & $7.58$    & $9.38$\\
    \hline
    Ours         & 141.07    & 30.84 & 2.88 &  $13.75$   & $2.97$  \\
    \hline\hline
    \end{tabular}   
\label{TAB4}
\end{table*}

\subsection{Ablation Study}
In this section, we conducted ablation studies for our CorrI2P on the KITTI dataset. \\

\noindent\textbf{Cross-attention fusion module}.
The cross-attention module can fuse the information from the image and point cloud with each other, which facilitates overlapping region detection, feature extraction, and correspondence estimation. 
Thus, we conducted an ablation study to identify its necessity. Similar to P2P registration \cite{PPFNET}, we trained a network without the cross attention fusion module and detection loss and used pair recall $R_\text{pair}$ and fragment recall $R_\text{frag}$ respectively defined in Eqs. (12) and (13) to evaluate the 2D-3D feature matching. 
Fig. \ref{FIG7A} shows the recall of feature matching through varying $\tau_1$ and $\tau_2$, 
convincingly demonstrating that the cross-attention fusion module is beneficial to feature matching and the establishment of the 2D-3D correspondence. \\


\noindent\textbf{Overlapping region detection}.
Our CorrI2P can detect the overlapping region of the image and point cloud, where dense 2D-3D correspondence is built. 
This can increase the inlier ratio of correspondence, thus achieving higher registration accuracy. To verify this, we also set baselines by employing different sampling strategies during the registration, including random and keypoint selection. We only used the descriptor loss to train the network for these experiments. For the random selection strategy, we kept the same number of sampled pixels and points, i.e., we randomly selected 2048 pixels and 8192 points from the image and point cloud, respectively, as well as their features to do the registration. For the keypoint selection strategy,  we imitated 2D3D-MatchNet, which uses SIFT and ISS to detect the key points from the image and point cloud to do registration, leading to sparse 2D-3D correspondence. We used 'Score' to represent our overlapping region detector because we used a confidence score to select the pixels and points. Besides, we also mixed these sampling strategies with ours, such as 'IMG Random' and 'PC Score'. The result is shown in Fig. \ref{FIG7B} and Table \ref{TAB2}. It can be seen that without our overlapping region detection, the registration accuracy would significantly decrease. However, only by removing the overlap detection for image, i.e., 'IMG Random + PC Score' or 'IMG SIFT + PC Score', would the accuracy decrease only a little while removing that for point cloud, i.e. 'PC Random' or 'PC ISS' the registration performance would decrease significantly. The points that can be projected onto the image only take up a small part of the whole point cloud, while the overlapping region on the image accounts for a large proportion as shown in Fig. \ref{overlap:gt:img}. \\

\noindent\textbf{3D point cloud density}. Considering that the density of 3D point clouds is a key factor in feature extraction, we carried out the ablation experiment on it. For a scene of the same size, we changed the density of the point cloud by downsampling different numbers of points. To keep the same receptive fields for point clouds with different point densities, we scaled the numbers of knn searching for different densities, i.e., $k=32$ for 40960 points, $k=16$ for 20480 points, and $k=8$ for 10240 and 5120 points. The result is shown in Table \ref{TAB3}, the registration accuracy decreases with reducing 3D point density because point cloud in low density would omit some structure information, and the features extracted would be less descriptive, resulting in wrong 2D-3D correspondence and thus low registration accuracy.

\subsection{Efficiency Analysis}
We evaluated the efficiency of our CorrI2P on the KITTI dataset. We used an NVIDIA Geforce RTX 3090 GPU for neural network inference and Intel(R) Xeon(R) Gold 6346 CPU for pose estimation. We fed data with a batch size of 8 to the neural network and got the average FLOPs, GPU memory usage, and inference time. The results are shown in Table \ref{TAB4}, where the classification-based methods, i.e., Grid Cls. and DeepI2P, require less GPU resource and are faster than ours during inference, because our method needs to perform feature extraction and overlapping region detection on both images and point clouds, i.e., using image and point cloud decoders to produce pixel-wise and point-wise features and scores, rather than only classifying the points of the point cloud. 
As for pose estimation, Grid Cls. + EPnP is the fastest because the image grid is 32$\times$32, resulting in a higher inlier ratio than ours, and the RANSAC only needs fewer iterations. DeepI2P is much slower than other methods because it needs 60-fold pose initialization before the optimization to enhance the robustness.

\section{Conclusion}
\label{sec:con}
We have presented a new learning-based framework named CorrI2P for 2D image-to-3D point cloud registration by estimating the dense correspondence between the two data modalities. 
Technically,  
we designed a symmetric overlapping region detector for both images and point clouds to estimate the overlapping regions, where dense 2D-3D correspondence is estimated based on their features. 
We demonstrated the significant advantages of our CorrI2P over state-of-the-art ones by conducting extensive experiments on the KITTI and NuScenes datasets, as well as comprehensive ablation studies. We believe our methods will benefit other tasks, such as distillation on image or point cloud and semantic segmentation for cross-modality data, which usually transform the different kinds of data to the same feature space.


\bibliographystyle{IEEEtran}
\bibliography{ref}

\vspace{-10mm}
\begin{IEEEbiography}[{\includegraphics[width=1in, height=1.25in, clip, keepaspectratio]{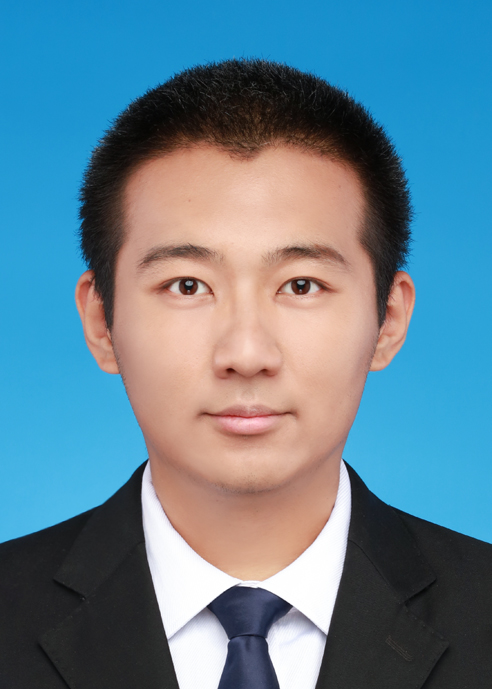}}]{Siyu Ren} received the B.S. degree in Optoelectronic Information Science and Engineering from Tianjin University, Tianjin, China, in 2018. 
He is currently pursuing the Ph.D. degree in Computer Science at the City University of Hong Kong and Optical Engineering at the Tianjin University. His research interests include deep learning and 3D point cloud processing.
\end{IEEEbiography}
\begin{IEEEbiography}[{\includegraphics[width=1in, height=1.25in, clip, keepaspectratio]{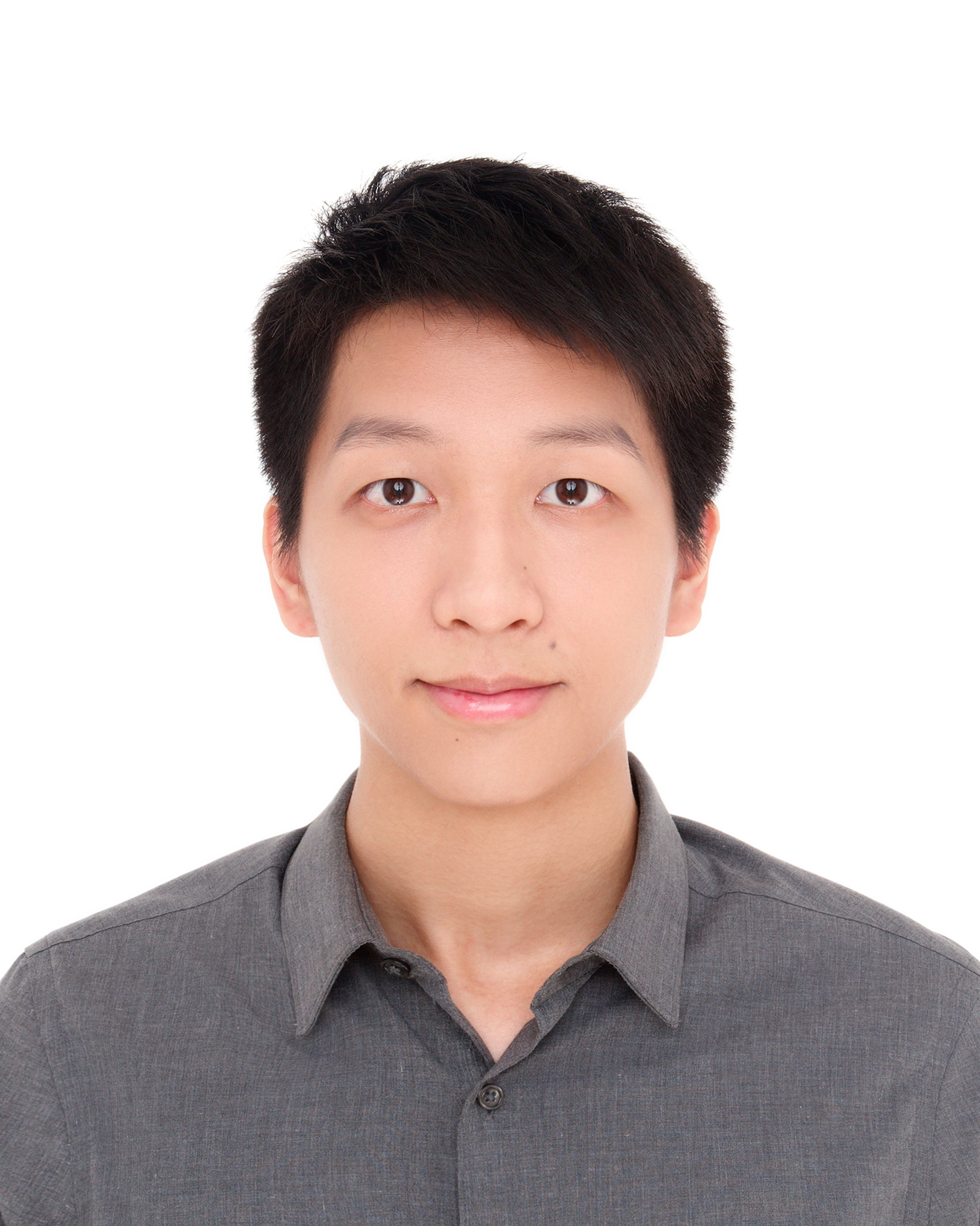}}]{Yiming Zeng} received the B.S. degree in automation from South China University of Technology, Guangzhou, China, in 2019. 
He is currently pursuing the Ph.D. degree in Computer Science at the City University of Hong Kong.  His research interests include deep learning and 3D point cloud processing.
\end{IEEEbiography}
\begin{IEEEbiography}[{\includegraphics[width=1in, height=1.25in, clip, keepaspectratio]{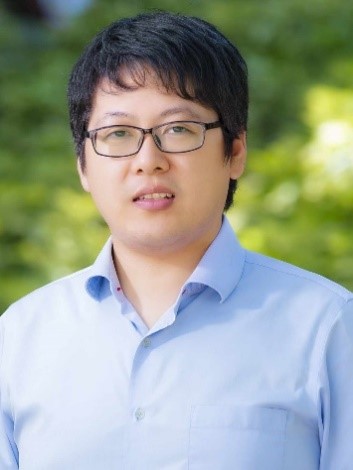}}]{Junhui Hou} (Senior Member)  is an Assistant Professor with the Department of Computer Science, City University of Hong Kong. He received the B.Eng. degree in information engineering (Talented Students Program) from the South China University of Technology, Guangzhou, China, in 2009, the M.Eng. degree in signal and information processing from Northwestern Polytechnical University, Xian, China, in 2012, and the Ph.D. degree in electrical and electronic engineering from the School of Electrical and Electronic Engineering, Nanyang Technological University, Singapore, in 2016. His research interests fall into the general areas of multimedia signal processing, such as image/video/3D geometry data representation, processing and analysis, graph-based clustering/classification, and data compression.

He received the Chinese Government Award for Outstanding Students Study Abroad from China Scholarship Council in 2015 and the Early Career Award (3/381) from the Hong Kong Research Grants Council in 2018. He is an elected member of IEEE MSA-TC, IEEE VSPC-TC, and IEEE MMSP-TC. He is currently an Associate Editor for IEEE Transactions on Image Processing, IEEE Transactions on Circuits and Systems for Video Technology, Signal Processing: Image Communication, and The Visual Computer. He also served as the Guest Editor for the IEEE Journal of Selected Topics in Applied Earth Observations and Remote Sensing and as an Area Chair of ACM MM’19-22, IEEE ICME’20, VCIP’20-22, ICIP’22, MMSP’22, and WACV’21.
\end{IEEEbiography}
\begin{IEEEbiography}[{\includegraphics[width=1in, height=1.25in, clip, keepaspectratio]{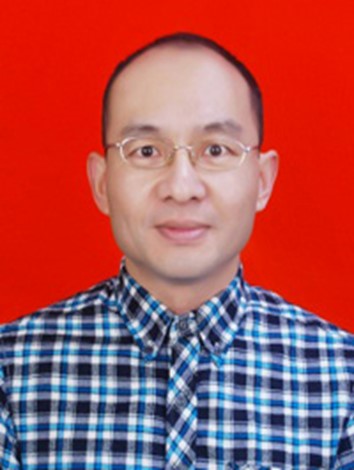}}]{Xiaodong Chen} received the Ph.D. degree in Optical Engineering at Tianjin University. He is a Professor with the School of Precision Instruments and Opto-Electronic Engineering, Tianjin University. He is the author of 2 books, more than 180 articles, and more than 7 inventions. His research interests include photoelectric detection technology and instrument, image processing and machine vision detection.
\end{IEEEbiography}

\end{document}